%% file: emnlp2023.tex
\pdfoutput=1
\documentclass[11pt]{article}
\usepackage{acl2023}
\usepackage{amsmath}

\usepackage{graphicx}
\usepackage{booktabs}
\usepackage{xcolor}
\usepackage[normalem]{ulem}
\usepackage{times}
\usepackage{latexsym}
\usepackage{pgfplots}
\usepackage{lipsum} 
\usepackage{multirow}
\usepackage{colortbl}
\usepackage{booktabs}
\usepackage{amssymb}
\usepackage{diagbox}
\usepackage{makecell}
\usepackage{array,booktabs,arydshln,xcolor}
\usepackage[T1]{fontenc}
\usepackage[utf8]{inputenc}

\usepackage{pifont}% http://ctan.org/pkg/pifont
\newcommand{\xmark}{\ding{55}}%

\usepackage{bbding}

\usepackage{microtype}

\usepackage{inconsolata}

\definecolor{model1}{RGB}{240, 29, 5}
\definecolor{model2}{RGB}{5, 95, 240}

\newcommand{\dataset}{\textsc{MisMatched}}

\title{A \textsc{MisMatched} Benchmark for Scientific Natural Language Inference}

\author{
    Firoz Shaik$^{\clubsuit}$ \mbox{     }\mbox{     } 
    Mobashir Sadat$^{\clubsuit}$ \mbox{     }\mbox{     } 
    Nikita Gautam$^{\blacklozenge}$ \mbox{     }\mbox{     } 
    Doina Caragea$^{\blacklozenge}$ \mbox{     }\mbox{     } 
    Cornelia Caragea$^{\clubsuit}$ \\
    \\
    Computer Science \\
    $^{\clubsuit}$University of Illinois Chicago \\
    $^{\blacklozenge}$Kansas State University \\
    {\texttt{\{fshaik8,msadat3,cornelia\}@uic.edu, \{ngautam,dcaragea\}@ksu.edu}}
}

\begin{document}
\maketitle
\begin{abstract}

Scientific Natural Language Inference (NLI) is the task of predicting the semantic relation between a pair of sentences extracted from research articles. Existing datasets for this task are derived from various computer science (CS) domains, whereas non-CS domains are completely ignored. In this paper, we introduce a novel evaluation benchmark for scientific NLI, called {\dataset}. The new {\dataset} benchmark covers three non-CS domains--\textsc{Psychology}, \textsc{Engineering}, and \textsc{Public Health}, and contains $2,700$ human annotated sentence pairs. We establish strong baselines on {\dataset} using both Pre-trained Small Language Models (SLMs) and Large Language Models (LLMs). Our best performing baseline shows a Macro F1 of only $78.17\%$ illustrating the substantial headroom for future improvements. In addition to introducing the {\dataset} benchmark, we show that incorporating sentence pairs having an \textit{implicit} scientific NLI relation between them in model training improves their performance on scientific NLI. We make our dataset and code publicly available on GitHub.\footnote{https://github.com/fshaik8/MisMatched} 
% \cornelia{\textsc{MSciNLI_{mm}??}}
%we perform a thorough analysis on incorporating sentence pairs having an \textit{implicit} scientific NLI relation in model training and find that they can aid improving the performance of scientific NLI models. We will make our code and data available on GitHub.

%the \textit{explicit} signal from linking phrases is essential for improving the model performance. We will make our code and data available on GitHub.

\end{abstract}

\input{Version_0.tex}

%1. SciBERT, BERT, XLNet (if time permits)
%2. Phi-3, Gemini/llama-3 (few-shot SciNLI, MSciNLI, MSciNLI+)
%3. Few-shot from Mismatched (4 from each domain separately)
%4, MNLI examples as few-shot --- contradiction to contrast, others remain same
%5. implicits multiple times
%6. 8 exemplars in the prompt from MSciNLI+ for a subset of models

\end{document}

%% file: Version_0.tex
\section{Introduction}
%\vspace{-2mm}
%scientific NLI is the task of ... it has been shown to improve the performance of downstram tasks.
%the task relies on linking phrases, i.e, relies on the relations being made explicit. 
%there can be domains where explicit relations are not in abundance. How do we explore scientific NLI for those domains?
%From discource coherence theory, we know that there can be implicit relations as well. However, the challenge is to detect them. 
%with the advancement of transfer learning, we can now train the models on one domain where we have a lot of explicit data and then assign labels to implicit examples for the target domain.
%

% Scientific Natural Language Inference (NLI) \cite{sadat-caragea-2022-scinli} classifies the semantic relation between a pair of sentences extracted from research articles into one of four classes---\textsc{Entailment}, \textsc{Reasoning}, \textsc{Contrasting}, and \textsc{Neutral}. This task is highly challenging for both Pre-trained Language Models (PLMs) and Large Language Models (LLMs). Therefore, scientific NLI is a suitable task to serve as a challenging benchmark for evaluating the Natural Language Understanding (NLU) of state-of-the-art models. In addition, research has shown that scientific NLI \cite{sadat2024mscinli} can aid in improving the performance of other downstream tasks such as topic classification and citation intent classification.

The task of Natural Language Inference (NLI) has received significant attention,  initially through several PASCAL %\footnote{https://k4all.org/project/25/} 
Recognising Textual Entailment (RTE) Challenges \cite{dagan06rte,haim2006second,giampiccolo2007third,bentivogli2009fifth} which focused on recognizing if two given sentences exhibit an entailment relationship. Subsequently, several NLI datasets \cite{bowman-etal-2015-large,williams-etal-2018-broad,nie-etal-2020-adversarial} have been introduced to facilitate progress on the NLI task. % through the means of Pre-trained Language Models (PLMs) and Large Language Models (LLMs). %Unlike RTE, in which sentence pairs are labeled as {\it entailment} or {\it non-entailment}, in the more recent NLI datasets, pairs are generally labeled according to three classes: {\it entailment}, {\it contradiction} and {\it neutral}. 
More recently, there has been an increasing interest in domain specific NLI tasks, including scientific NLI \cite{sadat-caragea-2022-scinli}.
The scientific NLI task classifies the semantic relation between a pair of sentences extracted from research articles into one of four classes---\textsc{Entailment}, \textsc{Reasoning}, \textsc{Contrasting}, and \textsc{Neutral}. This task is challenging for both Pre-trained Small Language Models (SLMs) and Large Language Models (LLMs) \cite{sadat-caragea-2024-mscinli}, making it suitable to serve as a challenging benchmark for evaluating the natural language understanding of state-of-the-art models. In addition,  \citet{sadat-caragea-2024-mscinli} have shown that scientific NLI can aid in improving the performance of other downstream tasks such as topic classification and citation intent classification.

%. Therefore, scientific NLI is a suitable task to serve as a challenging benchmark for evaluating the Natural Language Understanding (NLU) of state-of-the-art models. In addition, research has shown that scientific NLI \cite{sadat2024mscinli} can aid in improving the performance of other downstream tasks such as topic classification and citation intent classification.

To date, two datasets have been made available to facilitate research on scientific NLI---\textsc{SciNLI} \cite{sadat-caragea-2022-scinli}, and \textsc{MSciNLI} \cite{sadat-caragea-2024-mscinli}. \textsc{SciNLI} is derived from papers published in the ACL anthology, related to Natural Language Processing (NLP) and computational linguistics. To introduce diversity in scientific NLI, \textsc{MSciNLI} is constructed using sentence pairs extracted from five different scientific domains---\textsc{Hardware}, \textsc{Networks}, \textsc{Software \& its Engineering}, \textsc{Security \& Privacy}, and the \textsc{NeurIPS} conference which is related to machine learning. The training sets of these datasets are constructed using a distant supervision method that harnesses \textit{explicit} signals conveyed by various linking phrases. For example, if the second sentence in an adjacent sentence pair starts with ``However'' or ``In contrast,'' the sentence pair is labeled as \textsc{Contrasting}. The test and development sets of both \textsc{SciNLI} and \textsc{MSciNLI} are human annotated to ensure a realistic evaluation. 

%Both of these dataset are constructed using a distant supervision method based on linking phrases. For example, if the second sentence in an adjacent sentence pair starts with "However" or "In contrast," the sentence pair is labeled as \textsc{Contrasting}. Therefore, the construction process of both \textsc{SciNLI} and \textsc{MSciNLI} relies on sentence pairs where the scientific NLI relation between them is made \textit{explicit} using linking phrases. The automatically annotated sentence pairs with \textit{explicit} relations are directly included in the training sets. However, the test and development sets of both \textsc{SciNLI} and \textsc{MSciNLI} are human annotated. 

Despite the diversity introduced in \textsc{MSciNLI}, the domains covered by the existing scientific NLI datasets are still related to only computer science (CS), while non-CS domains are completely ignored. %Given the unique linguistic characteristics exhibited by text in different scientific domains,
%To capture the particularities of text from research articles in non-CS domains, 
% Therefore, it is necessary to enhance the coverage of scientific NLI benchmarks to non-CS domains which are more distant from the CS domains covered by \textsc{SciNLI} and \textsc{MSciNLI}. %Furthermore, the training sets of both datasets are composed of only sentence pairs which have an \textit{explicit} relation between them. However,  
%Furthermore, the distant supervision method used for constructing their training sets assume the availability of large number of sentence pairs where the semantic relations between them are made \textit{explicit} using linking phrases. Therefore, in a scenario where the \textit{explicit} relations between sentences are scarce, the usefulness of scientific NLI cannot be harnessed by relying on distant supervision based on linking phrases. 
Thus, in this paper, we propose a new evaluation benchmark for scientific NLI called {\dataset}, which contains sentence pairs collected from $3$ non-CS domains: \textsc{Psychology}, \textsc{Engineering}, and \textsc{Public Health}. We constructed {\dataset} as an out-of-domain (OOD) test-bed for scientific NLI models. That is, {\dataset} contains only development and test sets that are human annotated (of sizes $300$ and $2400$, respectively), without any training data. {\dataset} is designed as an out-of-domain (\textsc{OOD}) benchmark for evaluating the robustness of scientific NLI models, similar to the \textsc{mismatched} (\textsc{mm}) portion of \textsc{MNLI} \cite{williams-etal-2018-broad}. Like \textsc{MNLI}'s \textsc{mismatched} set, which uses unseen genres to test model generalization, {\dataset} is aimed at evaluating \textsc{OOD} robustness when models are trained on existing scientific NLI training sets. %The development and test sets of {\dataset} contain $300$ and $2,400$ sentence pairs respectively, balanced over both classes and domains. 

We establish strong baselines on {\dataset} by fine-tuning four SLMs---\textsc{BERT} \cite{devlin-etal-2019-bert}, \textsc{SciBERT} \cite{beltagy-etal-2019-scibert}, \textsc{RoBERTa} \cite{liu2019roberta} and \textsc{XLNet} \cite{yang2019xlnet}; and by prompting four LLMs---\textsc{Llama-2} \cite{touvron2023llama}, \textsc{Llama-3} \cite{grattafiori2024llama3}, \textsc{Mistral} \cite{jiang2023mistral7b} and \textsc{Phi-3} \cite{abdin2024phi} using the training sets from existing scientific NLI datasets. We find that our best performing SLM baseline with \textsc{SciBERT} and best performing LLM baseline with \textsc{Phi-3} show Macro F1 of only $78.17\%$ and $57.16\%$, respectively, illustrating the highly challenging nature of the {\dataset} set, and a significant amount of headroom for future improvements. In addition, given that {\em all} sentence pairs in the training sets of existing scientific NLI datasets are constructed using distant supervision that harnesses \textit{explicit} relations conveyed by various linking phrases, we analyze the impact on models' performance of sentence pairs which have an \textit{implicit} relation between them (i.e., sentence pairs which are adjacent in text and have a scientific NLI relation but the second sentence in the pair does not start with a linking phrase). We find that incorporating \textit{implicit} relations can indeed improve the performance of scientific NLI models. Our key contributions can be summarized as follows:

\begin{itemize}
    \item We introduce a novel {\dataset} benchmark which is more distant from computer science domains to further enhance the diversity of scientific NLI. 
    \item We establish strong baselines on {\dataset} using both SLMs and LLMs and show that it presents a challenging new benchmark for out-of-domain NLI evaluation.
    \item We incorporate \textit{implicit} relations in the scientific NLI model training, and show that %together with {\em explicit} relations, 
    they can improve the performance of scientific NLI models trained using only \textit{explicit} relations.
    
    %incorporate sentence pairs with implicit relations from these domains to adapt the baseline models trained only using explicit examples to further improve the performance on these baseline models. 
\end{itemize}

%we aim at improving the performance of scientific NLI models by utilizing sentence pairs with \textit{implicit} relations between them in three new 

%This task was proposed as a benchmark for evaluating Natural Language Understanding (NLU) of scientific text. 

\setlength\dashlinedash{0.2pt}
\setlength\dashlinegap{1.5pt}
\setlength\arrayrulewidth{0.3pt}
\begin{table*}[t]

\centering
\small

\scalebox{0.95}{
  \begin{tabular} { p{5em} p{17em}  p{17em} p{5em}}
    \toprule
%    \cline{1-2}
   \multicolumn{1}{l}{\rule{0pt}{1ex}\textbf{Domain}} & \multicolumn{1}{c}{\rule{0pt}{1ex}\textbf{First Sentence}} & \multicolumn{1}{c}{\textbf{Second Sentence}}&
   \multicolumn{1}{c}{\rule{0pt}{1ex}\textbf{Class}}\\
%   \cline{1-2}
    \hline%\rule{0pt}{3ex}
    \textsc{Engineering} & In previous studies, GBRS has acted as a guideline to improve energy use and indoor air quality. &  \sout{However,} the effectiveness of GBRS as applied to construction waste management has not been explored. &  \textsc{Contrasting}\\
    \hdashline
    \textsc{Public Health}  & For example, sewage-associated marker genes such as Bacteroides HF183 and HPyV, and enteric viruses such as human NoV are predominantly associated with human feces or sewage. & \sout{Therefore,} these marker genes can be used as a proxy to determine the risk associated with NoV and other enteric pathogens specific to sewage. & \textsc{Reasoning} \\
    \hdashline
     \textsc{Psychology}  & The presence of BED in one or both parents was associated with the emotional and behavioural development in offspring. & 
     \sout{Particularly,} the diagnosis of BED in both parents had a direct effect on infants' affective problems. &  \textsc{Entailment}\\
    \hdashline
     \textsc{Engineering} & This baffle geometry was tested for a well known seismic excitation (El Centro) and it was observed to effectively suppress free surface fluctuations and the slosh forces. & storage tank designers should ensure safe design margins and develop methodologies to overcome a wide range of possible scenarios. &  \textsc{Neutral}\\
    %\hdashline

    \bottomrule
    
  \end{tabular}}
  %\vspace{-2mm}
  \caption{\small Examples of sentence pairs from {\dataset}, extracted from different domains. The linking phrases at the beginning of the second sentence (strikethrough text in the table) are deleted after extracting the pairs and assigning the labels.}
  
  %\vspace{-2mm}
  
    \label{table:class_examples}
    %\vspace{-2mm}
\end{table*}

\section{Related Work}

%{\bf Natural Language Inference Datasets.}  %In general natural language inference, for a given premise-hypothesis pair, the task is to classify the pair according to a specific set of labels. 
%Many datasets have been created for general NLI.
Since the introduction of the NLI task \cite{dagan06rte}, numerous datasets have been introduced that include sentence pairs from the general domain. \textsc{RTE}  \cite{dagan06rte} is an early  dataset, which went through several iterations \cite{haim2006second,giampiccolo2007third,bentivogli2009fifth}. The \textsc{RTE} dataset contains premise-hypothesis pairs, which  are labeled as {\it entailment} or {\it non-entailment}. More recent datasets such as \textsc{SICK} \cite{marelli-etal-2014-sick}, \textsc{SNLI} \cite{bowman-etal-2015-large} and \textsc{MNLI}  \cite{williams-etal-2018-broad} contain sentence pairs that are classified as  {\it entailment}, {\it contradiction}, or {\it neutral}. The \textsc{SICK} dataset  \cite{marelli-etal-2014-sick} contains sentence pairs automatically extracted from paired image captions and video descriptions.  \textsc{SNLI} \cite{bowman-etal-2015-large}  contains premise-hypothesis pairs, where the premises are extracted from image captions, and the hypotheses are manually written by human annotators. %Despite its size, the \textsc{SNLI} dataset lacks diversity. To address this limitation, 
\textsc{MNLI} \cite{williams-etal-2018-broad} contains premise-hypothesis pairs where premises are
%The \textsc{MultiNLI} \cite{N18-1101} dataset was constructed to address this limitation. 
% The premises in \textsc{MNLI} are 
extracted from a variety of sources such as travel guides and face-to-face conversation,  while the hypotheses are manually written by human annotators, as in \textsc{SNLI}. 
%The \textsc{MultiNLI} dataset contains 433k premise-hypothesis pairs with large domain coverage. 
\textsc{ANLI} \cite{nie-etal-2020-adversarial}, another NLI dataset, was constructed in an adversarial fashion with human annotators in the loop who were instructed to write  sentence pairs for which the models make mistakes in their predictions. %\textsc{ANLI} contains more than 160k pairs. 

%created to increase the  difficulty of the prior datasets and also the robustness of the models. The dataset construction and model training were done iteratively with the goal of making the dataset harder and the model stronger at each iteration. To achieve this, human annotators wrote each hypothesis manually (having knowledge about the premise and label) using an adversarial approach,  which ended when the model failed to correctly predict the  label of the pair. The \textsc{ANLI} dataset contains more than 160k pairs. 

Several domain-specific NLI datasets have also been introduced. For example, \textsc{MedNLI} \cite{romanov-shivade-2018-lessons} was derived from medical records of patients with the premise-hypothesis pairs being annotated by experts (in the medical domain) as {\it entailment}, {\it contradiction}, or {\it neutral}. The NLI4CT dataset \cite{jullien2023nli4ct} contains premise-hypothesis pairs, in the form of clinical trial reports (CTR) and statements, labeled as {\it entailment} or {\it contradiction} by human annotators. NLI4CT-P (Perturbed) \cite{jullien2024semeval} is an extension of the original NLI4CT dataset \cite{jullien2023nli4ct} and was obtained by adding a contrast set derived from  perturbations to the original statements, to facilitate causal analyses.  

Most relevant to our work, \textsc{SciNLI} \cite{sadat-caragea-2022-scinli} is a scientific NLI dataset constructed from research articles published in the ACL Anthology, where  sentence pairs were extracted automatically from articles based on linking phrases, and classified into one of the following four classes: {\it entailment}, {\it reasoning}, {\it contrasting}, and {\it neutral} (with manual annotations only for the test and dev sets). \textsc{MSciNLI} \cite{sadat-caragea-2024-mscinli} is an extension of \textsc{SciNLI}, which was constructed from a larger variety of computer science research articles, e.g., \textsc{Hardware}, \textsc{Networks}, etc. and contains sentence pairs labeled also with the above four classes. To further diversify the datasets and study the transferability and robustness of the models for scientific NLI in out-of-distribution settings, in this paper, we introduce test/dev sets that cover articles from three non-computer science domains, specifically \textsc{Psychology}, \textsc{Engineering}, and \textsc{Public Health}. %Our dev/test sets contain 100 and 800 manually annotated premise-hypothesis pairs for each domain, respectively, for a total of 2,700 pairs. 

A comparison of all the datasets reviewed here is shown in Appendix \ref{appendix-rel-work}.

% \noindent
% Approaches 

% \textsc{MULTIVERS} \cite{wadden2022multivers}

% \cite{wu-etal-2023-characterizing}

% \cite{gema2024edinburgh}

\setlength\dashlinedash{0.2pt}
\setlength\dashlinegap{1.5pt}
\setlength\arrayrulewidth{0.3pt}

\begin{table*}[t]
\centering
\small

\scalebox{1}{
  \begin{tabular}{ l r r r r r r r r r r }
    \toprule
      &  \multicolumn{3}{c}{\bf \#Examples} & \multicolumn{2}{c}{\bf \#Words} & \multicolumn{2}{c}{\bf `S' parser} & \multicolumn{1}{c}{\bf Word} \\
       \cmidrule(lr){2-4}  \cmidrule(lr){5-6}  \cmidrule(lr){7-8}
      %\cline{2-4}
   {\bf Dataset } & {\bf Train}  & {\bf Dev}   & {\bf Test} & {\bf Prem.} & {\bf Hyp.} & {\bf Prem.} & {\bf Hyp.} & {\bf Overlap} & {\bf \#Domains} & {\bf Agrmt.}  \\ %\cline{2-4}
   \midrule
   \textsc{SciNLI (ACL)} & 101,412 & 2,000 & 4,000 & 27.38 & 25.93 & 96.8\% & 96.7\% & 30.06\% & 1 & 85.8\%\\  %\hline
\hdashline
\textsc{MSciNLI} & 127,320 & 1,000 & 4,000 & 26.84 & 25.85 & 94.4\% & 94.3\% & 30.29\% & 5 & 88.0\%\\
\hdashline
{\bf \dataset}  & - & 300 & 2400 & 26.65 & 25.75 & 96.8\% & 98.2\% & 31.27\% & 3 & 85.7\% \\
$\diamondsuit$ \mbox{{\color{white}{x}}}\textsc{Public Health} & - & 100 & 800 & 27.42 & 27.22 & 98.4\% & 97.8\% & 31.19\% & 1 & 84.3\% \\
$\diamondsuit$  \mbox{{\color{white}{x}}}\textsc{Psychology} & - & 100 & 800 & 25.95 & 25.59 & 94.1\% & 97.7\% & 31.01\% & 1 & 88.3\% \\
$\diamondsuit$  \mbox{{\color{white}{x}}}\textsc{Engineering} & - & 100 & 800 & 26.59 & 24.45 & 97.8\% & 98.8\% & 31.59\% & 1 & 85.6\% \\
    \bottomrule
  \end{tabular}}
  %\vspace{-3.5mm}
  \caption{\small Comparison of the key statistics of the {\dataset} set with \textsc{MSciNLI} and \textsc{SciNLI}.}
  \vspace{-4mm}
    \label{table:data_stat}
\end{table*}

\section{The {\dataset} Benchmark}% for Scientific NLI}
\vspace{-2mm}
In this section, we describe our proposed {\dataset} benchmark for scientific NLI. Specifically, we outline the data sources for deriving {\dataset}, the construction process and the key statistics. Table \ref{table:class_examples} shows examples of sentence pairs from different domains and classes in our %resulting 
dataset.

\subsection{Data Sources} 
Our {\dataset} benchmark is composed of three domains---\textsc{Psychology}, \textsc{Engineering}, and \textsc{Public Health}. We selected these domains to extend scientific NLI beyond existing computer science focused datasets. While \textsc{SciNLI} \cite{sadat-caragea-2022-scinli} covers computational linguistics and \textsc{MSciNLI} \cite{sadat-caragea-2024-mscinli} encompasses CS domains (Hardware, Networks, Software \& Engineering, Security \& Privacy, and NeurIPS), our new domains represent diverse non-CS scientific areas with broad real-world applications. %Domain selection was also guided by our team's expertise to ensure rigorous annotation quality validation. 
The data sources for each of these domains are described below.

\paragraph{\textsc{Psychology.}} %The sentence pairs for the \textsc{Psychology} are extracted solely from abstracts of papers from the relevant domain in WoS. Particularly, we 
\citet{kowsari2017hdltex} constructed a dataset for topic classification of scientific papers. The dataset contains abstracts from Web of Science (WoS) papers, which belong to 7 scientific domains, including the Psychology and Engineering domains. WoS is a database that indexes global scholarly literature across sciences from various journals and academic events. We extract sentence pairs from papers in the \textsc{Psychology} domain for our {\dataset} set. % from the papers classified to the \textsc{Psychology} domain in this topic classification dataset. 

\paragraph{\textsc{Engineering.}} For the \textsc{Engineering} domain, we also utilize a subset of WoS papers  from the topic classification dataset introduced by \citet{kowsari2017hdltex}. Specifically, we extract sentence pairs from   ``Civil Engineering'', ``Electrical Engineering'' \& ``Mechanical Engineering'' papers. 

\paragraph{\textsc{Public Health.}} %The sentence pairs for the \textsc{Public Health} domain are extracted from various research papers from the National Library of Medicine, the Web of Science (WoS) platform,  Centers for Disease Control and Prevention (CDC), and Environmental Protection Agency (EPA). Note that for the papers from WoS, we only use the open source abstracts to construct our dataset. \mobashir{need more details about the collection of papers}.

% {\color{blue}Three sources were used to extract sentence pairs for the Public Health domain dataset. Using twenty-five keywords related to marine water characteristics and the health risks of divers and swimmers (such as coastal water pollution and beach water contamination), almost 100k abstracts from the Web Of Science (WoS) were crawled. Web of Science is a database that indexes global scholarly literature across sciences from various journals and academic events. Next, the National Library of Medicine (NLM) was used to collect 200 additional abstracts for articles specific to water diving. Finally, 153 full-text scholarly articles and reports related to the Centers for Disease Control and Prevention (CDC) and the U.S. Environmental Protection Agency (EPA) were collected using a manual PubMed search of biomedical literature from MEDLINE, life science journals, and online books. During the initial pre-processing of the collected dataset, non-English and duplicate texts were removed. Only the open-source abstracts and full texts were used to construct the dataset.}

Three sources are used to extract sentence pairs for the \textsc{Public Health} domain. Using twenty-five keywords related to marine water characteristics and the health risks of divers and swimmers (such as coastal water pollution and beach water contamination), we crawled about 100k abstracts from WoS. Next, the National Library of Medicine (NLM) was used to collect 200 additional abstracts for articles specific to water diving. Finally, 153 full-text scholarly articles and reports related to the Centers for Disease Control and Prevention (CDC) and the U.S. Environmental Protection Agency (EPA) were collected using a manual PubMed search of biomedical literature from MEDLINE, life science journals, and online books. During the initial pre-processing of the collected papers, non-English and duplicate texts were removed. Only the open-source abstracts and full texts were used to construct our dataset.

\subsection{Dataset Construction} To construct our {\dataset} set, we follow a  procedure similar to that employed for creating the test and development sets of \textsc{SciNLI} and \textsc{MSciNLI}. In phase 1, we automatically extract and annotate sentence pairs using the distant supervision method proposed by \citet{sadat-caragea-2022-scinli}. In phase 2, we employ human annotators to curate the final test and development sets.

\paragraph{Phase 1: Automatic Data Extraction and Annotation.} %As the first step of our dataset construction process, we extract the sentence pairs from the aforementioned data sources using the distant supervision method proposed by \citet{sadat-caragea-2022-scinli}. 
For the \textsc{Entailment}, \textsc{Contrasting} and \textsc{Reasoning} classes, we automatically extract adjacent sentence pairs where the second sentence starts with a linking phrase indicative of these relations.  We then remove the linking phrase from the second sentence, and assign the label based on the semantic relation indicated by the linking phrase (as shown in Table \ref{table:class_examples}). For example, if the second sentence starts with ``Therefore'' or ``As a result,'' we extract and annotate the sentence pair with the \textsc{Reasoning} label. The mapping of linking phrases to labels can be seen in Appendix \ref{appendix:linking_phrases}.

For the \textsc{Neutral} class, we randomly pair two non-adjacent sentences from the same paper using 3 strategies: a) \textsc{BothRand}: two random sentences which do not contain any linking phrases are paired; b) \textsc{FirstRand}: a random sentence is paired with a second sentence from the other three classes; c) \textsc{SecondRand}: a random sentence is paired with a first sentence from the other three classes.

\vspace{-1mm}
\paragraph{Phase 2: Human Annotation.} To construct the final test and development sets, we hire annotators via a crowd-sourcing platform called COGITO.\footnote{https://www.cogitotech.com/} Note that separate annotators are hired for the three domains to ensure that the annotators have the  background knowledge and expertise necessary to understand  domain-specific sentences. More details on annotators (e.g., pilot batch completion, pay, etc.) are available in Appendix \ref{appendix: annotator_details}.

%The annotators are provided with two sentences from each pair and they are instructed to annotate the label based on only the context available in the two sentences. For each domain, we perform three pilot batches of annotations to train the annotators. After each pilot batch, we provide the annotators with detailed feedback. The annotators demonstrated a solid understanding of the scientific NLI task after completing three pilot batches, and we proceeded with the final annotations for constructing the {\dataset} set. 

We perform the human annotations in an iterative fashion. In all iterations (except last), we randomly sample a balanced (over classes) subset of sentence pairs and ask three expert annotators to assign the label based only on the context available in the two sentences in each pair. Based on the consensus of the annotators, we assign a gold label to each example. The examples for which the gold label matches with the automatically assigned label based on distant supervision are included in the {\dataset} set, and the rest are discarded. For each domain, we continue the iterations 
% of sampling, annotating, and selecting examples 
until we have at least $225$ examples from each of the non-\textsc{Neutral} classes. % in the {\dataset} set. 
%cornelia-comment
For the \textsc{Neutral} class, we notice a lower agreement rate between the gold label and the automatically assigned label in all domains. Thus, for each domain, we perform a last iteration with all sentence pairs sampled from the \textsc{Neutral} class to obtain (at least) $225$ examples %(same as the other classes) 
where the automatically assigned \textsc{Neutral} label matches with the human annotated gold label. The distribution of the automatically assigned labels is not made available to the annotators for any batch.

\setlength\dashlinedash{0.2pt}
\setlength\dashlinegap{1.5pt}
\setlength\arrayrulewidth{0.3pt}
\begin{table*}[t]
\centering
\small

  \begin{tabular}{p{45em}}
    \toprule
  <human>: Consider the following two sentences:\newline
   {\color{blue} Sentence1: <sentence1>}\newline
 {\color{orange} Sentence2: <sentence2>}\newline
Based  only on the information available in these two sentences, which of the following options is true?\newline
a. {\color{blue} Sentence1} generalizes, specifies or has an equivalent meaning with  {\color{orange} Sentence2}.\newline
b. {\color{blue} Sentence1} presents the reason, cause, or condition for the result or conclusion made {\color{orange} Sentence2}.\newline
c. {\color{orange} Sentence2} mentions a comparison, criticism, juxtaposition, or a limitation of something said in {\color{blue} Sentence1}.\newline
d. {\color{blue} Sentence1} and {\color{orange} Sentence2} are independent.\newline
<bot>:\\
    %{\bf \textsc{Prompt - 5}} & $0.70$ \\
    \bottomrule
  \end{tabular}

  \caption{\small Prompt template used for LLMs. Here, <X> indicates a placeholder X, which is replaced in the actual prompt.}

    \label{table:prompts}
    \vspace{-2mm}
\end{table*}

%We set a target size of $225$ examples in each class for each domain in {\dataset}.

%For all domains, we notice a lower agreement rate between the gold label and the automatically assigned label for the \textsc{Neutral} class. Therefore, in the last iteration for all domains, we only annotate a batch sampled from the automatically annotated \textsc{Neutral} class. 

%For each domain, we continue the iterations of sampling, annotating and selected examples for the {\dataset} set until there are at least $225$ examples from each of the four classes in the {\dataset} set.

%At each iteration, we randomly sample a balanced (over the classes) subset of sentence pairs and ask three expert annotators to assign the label. Based on the consensus of the annotators, we assign a gold label to each example. The examples for which the gold label matches with the automatically assigned label based on distant supervision are included in the {\dataset} set, and the rest are discarded. For each domain, we continue the iterations of sampling, annotating and selected examples for the {\dataset} set until there are at least $225$ examples from each of the four classes in the {\dataset} set. 
In total, we annotate $3,253$ sentence pairs, among which $2,791$ have an agreement between the gold label and the automatically assigned label. The annotators showed a Fleiss-$\kappa$ score of $0.72$ among them (see Appendix \ref{appendix: annotator_details} for domain-wise breakdown). The domain-wise agreement rates between the gold label and the automatically assigned label can be seen in Table \ref{table:data_stat}. We report the class-wise agreement rates in Appendix \ref{appendix:class_wise_agreement}.

\vspace{-1mm}
\paragraph{Data Balancing.} To ensure an equal representation of the classes and the domains, we randomly downsample all classes across domains to $225$ (our domain-wise target size for each class). %(this is the minimum support of a class over all domains). 
That is, the resulting {\dataset} set contains $2,700$ examples in total, uniformly distributed over the three domains ($900$ in each domain).

\vspace{-1mm}
\paragraph{Data Split.} %Similar to both \textsc{SciNLI} and \textsc{MSciNLI}, 
We split the {\dataset} set into test and dev sets at the paper level to prevent data leakage. Specifically, we randomly split the papers in each domain ensuring that there are at least $800$ and $100$ examples in the test and dev sets, respectively, with both sets being balanced over classes.

\subsection{Data Statistics} 

We report the key statistics of our {\dataset} set in Table \ref{table:data_stat}. As we can see, the per-domain test size of {\dataset} is the same as that of \textsc{MSciNLI} (both 800). While the per-domain dev size of {\dataset} is smaller compared with \textsc{MSciNLI}, it still contains a satisfactory number of examples to be able to perform validation and hyper-parameter tuning. We can also see that the average number of words in the sentences in {\dataset} is similar to that of the existing datasets. In addition, for both sentences, the percentage of sentences that have an ``S'' root according to the Stanford PCFG Parser (3.5.2) \cite{klein-manning-2003-accurate} is over $96\%$. This indicates that the vast majority of sentences in our dataset are syntactically complete. We can also see that the percentage of words that overlap between the two sentences is low, similar to the existing scientific NLI datasets. %Furthermore, the agreement rates

%Table \ref{table:data_stat} shows that the word overlap percentage between the premise and hypothesis is slightly higher in {\dataset} than \textsc{SciNLI} and \textsc{MSciNLI}. However, it is still only $31.27\%$ overall indicating a low overlap.

%This is because, unlike these datasets, most sentence pairs in {\dataset} is extracted from abstract of a paper. 

%comparison with SciNLI and MSciNLI
%size
%sentence lengths
%only abstracts
%S parser
%agreement

\begin{table*}[t]
\centering
\small
% \addlinespace

\scalebox{0.90}{
  \begin{tabular}{l c c c c}
    \toprule
%      &  \multicolumn{2}{c}{\bf SICK} & {\bf SciTail} & SNLI & SciNLI\\%\cline{2-4}
{\bf \textsc{Model}} &  {\bf \textsc{Psychology}} &  {\bf \textsc{Engineering}} & {\bf \textsc{Public Health}} & {\bf \textsc{Overall}}\\
    \midrule
    \textsc{BERT}$_{SciNLI}$ & $68.59 \pm 2.8$ & $69.26 \pm 2.3$ & $66.57 \pm 2.6$ & $68.16 \pm 2.5$ \\
    \textsc{BERT}$_{MSciNLI}$ & $68.00 \pm 1.4$ & $69.23 \pm 2.1$ & $66.34 \pm 1.2$ & $67.89 \pm 1.2$ \\
    \textsc{BERT}$_{MSciNLI+}$ & $71.16 \pm 0.9$ & $73.52 \pm 0.1$ & $69.47 \pm 1.3$ & $71.41 \pm 0.6$ \\
    \hdashline
    \textsc{SciBERT}$_{SciNLI}$ & $76.24 \pm 1.5$ & $74.36 \pm 1.4$ & $78.14 \pm 2.0$ & $76.27 \pm 1.6$ \\
    \textsc{SciBERT}$_{MSciNLI}$ & $76.98 \pm 1.2$ & $76.56 \pm 0.8$ & $77.97 \pm 0.8$ & $77.66 \pm 0.8$ \\
    \textsc{SciBERT}$_{MSciNLI+}$ & ${\bf 79.18 \pm 0.4}$ & $76.50 \pm 0.8$ & ${\bf 78.79 \pm 0.3}$ & ${\bf 78.17 \pm 0.2}$ \\
    \hdashline
    % \textsc{RoBERTa}$_{SciNLI}$ & $75.40 \pm 0.5$ & $75.38 \pm 0.6$ & $76.09 \pm 2.0$ & $75.64 \pm 0.9$\\
    % %\hdashline
    % \textsc{RoBERTa}$_{MSciNLI}$ & $74.87 \pm 0.5$ & $75.53 \pm 1.1$ & $75.18 \pm 0.5$ & $75.23 \pm 0.6$\\
    % %\hdashline
    % \textsc{RoBERTa}$_{MSciNLI+}$ & $77.28 \pm 0.2$ & $76.71 \pm 0.3$ & $77.43 \pm 0.2$ & $77.52 \pm 0.4$\\
    \textsc{RoBERTa}$_{SciNLI}$ & $75.76 \pm 0.1$ & $75.12 \pm 0.7$ & $75.34 \pm 1.5$ & $75.43 \pm 0.5$\\
    %\hdashline
    \textsc{RoBERTa}$_{MSciNLI}$ & $75.05 \pm 1.2$ & $76.07 \pm 0.8$ & $74.89 \pm 1.1$ & $75.38 \pm 1.0$\\
    %\hdashline
    \textsc{RoBERTa}$_{MSciNLI+}$ & $77.91 \pm 0.3$ & ${\bf 77.63 \pm 0.3}$ & ${\bf78.79 \pm 1.0}$ & $78.11 \pm 0.3$\\
    \hdashline
    \textsc{XLNet}$_{SciNLI}$ & $73.61 \pm 0.8$ & $72.61 \pm 0.7$ & $73.23 \pm 2.0$ & $73.19 \pm 1.0$ \\
    %\hdashline
    \textsc{XLNet}$_{MSciNLI}$ & $73.24 \pm 2.2$ & $74.31 \pm 1.0$ & $73.19 \pm 0.4$ & $73.60 \pm 1.2$\\
    %\hdashline
    \textsc{XLNet}$_{MSciNLI+}$ & $76.40 \pm 1.0$ & $75.44 \pm 2.1$ & $75.54 \pm 0.9$ & $76.49 \pm 1.3$\\
    
    %\hdashline
   % \textsc{RoBERTa}$_{MSciNLI+ (S)}$ & ${\bf 77.79^{\$} \pm 0.2}$ & $\underline {75.45 \pm 1.5}$ & ${\bf 77.10^{\#} \pm 0.7}$ & ${\bf 77.71^{\$} \pm 0.2}$\\
    %\mbox{{\color{white}{xxxxxxxxxx}}+}\textsc{Implicit}\\
    \midrule
    % \textsc{Phi-3}$_{zs}$ & $37.22$ & $38.94$ & $36.94$ & $37.81$\\
    % \textsc{Phi-3}$_{zs}$ & $40.94$ & $45.68$ & $41.79$ & $42.94$\\
    % \textsc{Phi-3}$_{fs - SciNLI}$ & $45.16 \pm 2.4$ & $41.34 \pm 1.6$ & $37.44 \pm 2.6$ & $41.42 \pm 2.1$\\
    % \textsc{Phi-3}$_{fs - MSciNLI}$ & $45.16 \pm 2.3$ & $46.16 \pm 2.1$ & $39.98 \pm 2.5$ & $43.85 \pm 2.2$\\
    % \textsc{Phi-3}$_{fs - MSciNLI+}$ & $44.52 \pm 2.4$ & $43.74 \pm 4.7$ & $38.81 \pm 3.6$ & $42.40 \pm 3.5$\\
    % \hdashline

    \textsc{Phi-3}$_{zs}$ & $55.38\pm0.00$ & $53.15\pm0.00$ & $49.31\pm0.00$ & $52.95\pm0.00$\\
    \textsc{Phi-3}$_{fs - SciNLI}$ & $57.98 \pm 1.31$ & $55.46 \pm 1.02$ & $53.53 \pm 0.77$ & $55.84 \pm 0.98$\\
    \textsc{Phi-3}$_{fs - MSciNLI}$ & $58.64 \pm 1.11$ & ${\bf 56.76 \pm 0.57}$ & ${\bf 55.68 \pm 0.25}$ & ${\bf 57.16 \pm 0.59}$\\
    \textsc{Phi-3}$_{fs - MSciNLI+}$ & ${\bf 59.02 \pm 0.34}$ & $55.53 \pm 0.80$ & $55.54 \pm 0.93$ & $56.87 \pm 0.25$\\
    \hdashline
    
    \textsc{Llama-2}$_{zs}$ & $26.37\pm0.00$ & $32.71\pm0.00$ & $27.25\pm0.00$ & $28.98\pm0.00$\\
    %\hdashline
    \textsc{Llama-2}$_{fs - SciNLI}$ & $43.92 \pm 0.93$ & $49.11 \pm 1.54$ & $45.09 \pm 2.84$ & $46.24 \pm 1.71$\\
    %\hdashline
    \textsc{Llama-2}$_{fs - MSciNLI}$ & $44.83 \pm 2.75$ & $50.26 \pm 1.63$ & $45.45 \pm 1.33$ & $47.09 \pm 1.88$\\
    \textsc{Llama-2}$_{fs - MSciNLI+}$ & $43.54 \pm 2.00$ & $49.05 \pm 1.56$ & $44.03 \pm 2.04$ & $45.79 \pm 1.84$\\
    \hdashline

    \textsc{Llama-3}$_{zs}$ & $33.67\pm0.00$ & $37.00\pm0.00$ & $30.87\pm0.00$ & $33.95\pm0.00$\\
    %\hdashline
    \textsc{Llama-3}$_{fs - SciNLI}$ & $51.18 \pm 1.11$ & $46.88 \pm 0.48$ & $45.68 \pm 2.26$ & $48.19 \pm 1.03$\\
    %\hdashline
    \textsc{Llama-3}$_{fs - MSciNLI}$ & $52.66 \pm 1.15$ & $47.54 \pm 0.39$ & $45.85 \pm 0.72$ & $48.94 \pm 0.10$\\
    \textsc{Llama-3}$_{fs - MSciNLI+}$ & $53.92 \pm 1.01$ & $50.18 \pm 1.10$ & $48.13 \pm 0.62$ & $51.02 \pm 0.51$\\
    \hdashline

    \textsc{Mistral}$_{zs}$ & $31.14\pm0.00$ & $34.70\pm0.00$ & $25.85\pm0.00$ & $30.63\pm0.00$\\
    %\hdashline
    \textsc{Mistral}$_{fs - SciNLI}$ & $44.26 \pm 2.69$ & $44.58 \pm 2.59$ & $39.79 \pm 3.87$ & $43.02 \pm 3.03$\\
    %\hdashline
    \textsc{Mistral}$_{fs - MSciNLI}$ & $47.04 \pm 1.82$ & $47.12 \pm 2.82$ & $43.68 \pm 2.68$ & $46.09 \pm 2.40$\\
    \textsc{Mistral}$_{fs - MSciNLI+}$ & $44.73 \pm 0.37$ & $45.46 \pm 0.71$ & $41.75 \pm 2.14$ & $44.09 \pm 0.87$\\
    % \textsc{Llama-2}$_{MSciNLI+}$ & ${\bf 77.79^{\$} \pm 0.2}$ & $\underline {75.45 \pm 1.5}$ & ${\bf 77.10^{\#} \pm 0.7}$ & ${\bf 77.71^{\$} \pm 0.2}$\\
    % \hdashline
    % %\textsc{Llama-2}$_{MSciNLI+}$ & ${\bf 77.79^{\$} \pm 0.2}$ & $\underline {75.45 \pm 1.5}$ & ${\bf 77.10^{\#} \pm 0.7}$ & ${\bf 77.71^{\$} \pm 0.2}$\\
    % %\hdashline
    % \textsc{Llama-2}$_{MSciNLI+ (S)}$ & ${\bf 77.79^{\$} \pm 0.2}$ & $\underline {75.45 \pm 1.5}$ & ${\bf 77.10^{\#} \pm 0.7}$ & ${\bf 77.71^{\$} \pm 0.2}$\\
    %\mbox{{\color{white}{xxxxxxx}}+}\textsc{Implicit}\\

    \midrule
    
    \textsc{GPT-4o}$_{zs}$ & $52.42\pm0.00$ & $50.12\pm0.00$ & $47.26\pm0.00$ & $50.26\pm0.00$\\
    %\hdashline
    \textsc{GPT-4o}$_{fs - SciNLI}$ & $63.33 \pm 1.52$ & $61.34 \pm 0.46$ & $61.62 \pm 0.50$ & $62.29 \pm 0.51$\\
    %\hdashline
    \textsc{GPT-4o}$_{fs - MSciNLI}$ & $62.65 \pm 2.31$ & $57.94 \pm 1.84$ & $58.61 \pm 0.72$ & $59.96 \pm 1.62$\\
    \textsc{GPT-4o}$_{fs - MSciNLI+}$ & $63.62 \pm 1.57$ & $61.06 \pm 0.83$ & ${\bf 62.96} \pm 1.22$ & $62.73 \pm 0.98$\\
    \hdashline

    \textsc{GEMINI-1.5-PRO}$_{zs}$ & $54.28\pm0.00$ & $58.49\pm0.00$ & $51.59\pm0.00$ & $55.55\pm0.00$\\
    %\hdashline
    \textsc{GEMINI-1.5-PRO}$_{fs - SciNLI}$ & $63.50 \pm 1.92$ & $61.94 \pm 1.41$ & $62.69 \pm 1.03$ & $62.78 \pm 1.41$\\
    %\hdashline
    \textsc{GEMINI-1.5-PRO}$_{fs - MSciNLI}$ & $63.09 \pm 0.86$ & $61.74 \pm 0.91$ & $62.53 \pm 0.59$ & $62.51 \pm 0.54$\\
    \textsc{GEMINI-1.5-PRO}$_{fs - MSciNLI+}$ & ${\bf 63.68} \pm 1.70$ & ${\bf 62.57} \pm 2.00$ & $62.51 \pm 1.20$ & ${\bf 62.95} \pm 1.50$\\
    
    \midrule

  \end{tabular}}
\vspace{-2mm}
   \caption{\small Macro F1 scores (\%) of the SLM and LLM baselines on different domains. Here, the subscript with the SLMs denotes the dataset used for fine-tuning the model. A subscript of $zs$ with LLMs indicates zero-shot setting, and $fs-X$ indicates few-shot setting with four exemplars (one per class) from dataset $X$. Best scores within SLM, Open-Source LLM, and Proprietary LLM baselines per domain and overall are shown in {\bf bold}.
   %Here, \textsc{SWE}: Software \& its Engineering and \textsc{Security}: Security \& Privacy. $^\#$ and $^\$$ indicate statistically significant improvement by \textsc{RoBERTa} over \textsc{XLNet} and over both \textsc{SciBERT} and \textsc{XLNet}, respectively according to a paired t-test with $p < 0.05$. Best performance is shown in {\bf bold}, and second best is {\underline{underlined}}. 
  }
  
    \label{table:baseline_results}
 \vspace{-3mm}
\end{table*}

\section{Baselines}
% \vspace{-2mm}
Since {\dataset} only consists of dev and test, we use the training sets of \textsc{SciNLI} and \textsc{MSciNLI}, containing $101K$ and $127K$ sentence pairs, respectively, to establish the SLM and LLM baselines. Our implementation details are in Appendix \ref{appendix:implementation_details}.

\subsection{Models}

\paragraph{SLM Baselines.} %For our PLM baselines, we fine-tune the base variants of \textsc{BERT} \cite{devlin-etal-2019-bert}, \textsc{SciBERT} \cite{beltagy-etal-2019-scibert}, \textsc{RoBERTa} \cite{liu2019roberta} and \textsc{XLNet} \cite{yang2019xlnet} using the training sets of \textsc{SciNLI}, \textsc{MSciNLI}, and a combination of \textsc{SciNLI} and \textsc{MSciNLI}, denoted as \textsc{MSciNLI+}.

%For our PLM baselines, 
We fine-tune the base variants of \textsc{BERT} \cite{devlin-etal-2019-bert}, \textsc{SciBERT} \cite{beltagy-etal-2019-scibert}, \textsc{RoBERTa} \cite{liu2019roberta} and \textsc{XLNet} \cite{yang2019xlnet} as our SLM baselines, using the training sets of \textsc{SciNLI}, \textsc{MSciNLI}, and their combination denoted as \textsc{MSciNLI+}.

%a combination of \textsc{SciNLI} and \textsc{MSciNLI}, denoted as \textsc{MSciNLI+}.

\paragraph{LLM Baselines.} Our selection of LLMs focused on popular, instruction-tuned models representing recent advancements suitable for prompt-based NLI and reproducible research. We experiment with several open-source LLMs, including the \textit{Llama-2-13b-chat-hf} variant of \textsc{Llama-2} \cite{touvron2023llama}, \textit{Llama-3.1-8B-Instruct} variant of \textsc{Llama-3} \cite{grattafiori2024llama3}, \textit{Mistral-7B-Instruct-v0.3} variant of \textsc{Mistral} \cite{jiang2023mistral7b} and \textit{Phi-3-medium-128k-instruct} (containing $14$B parameters) variant of \textsc{Phi-3} \cite{abdin2024phi}. Furthermore, to benchmark against leading proprietary models, we include \textsc{GPT-4o} \cite{hurst2024gpt4o} and \textsc{Gemini-1.5-Pro} \cite{georgiev2024gemini}. All LLMs are evaluated in both zero-shot and few-shot settings.  We use the best performing prompt constructed by (\citet{sadat-caragea-2024-mscinli}) for \textsc{MSciNLI}, shown in Table \ref{table:prompts}.
% In particular, we use a prompt where the two sentences in each sentence pair are shown to the LLM and then a multiple-choice question is asked where the class definitions of the four classes are used as the choices. 
In the zero-shot setting, no exemplars are provided to the model. In the few-shot setting, we prepend four exemplars in the prompt (one per class) to harness the LLMs' in-context learning ability.

%We perform three experiments for each LLM in the few-shot setting where we use two separate sets of randomly sampled exemplars from \textsc{SciNLI} and \textsc{MSciNLI} training sets. 

%Except the experiments with \textsc{Llama-2} in the zero-shot setting,
\subsection{Results \& Discussion}
\label{sec:results}
We report the domain-wise and overall Macro F1 of the LLMs in the zero-shot setting from a single run, since we use greedy decoding and therefore, there is no randomness involved. For all other experiments (with both SLMs and LLMs), we report the average and the standard deviations of the Macro F1 scores from three separate runs. Specifically, for the few-shot LLMs, we perform $3$ runs with $3$ randomly sampled sets of 4 exemplars from each \textsc{SciNLI}, \textsc{MSciNLI}, and \textsc{MSciNLI+} following the procedure detailed in Appendix \ref{appendix: Few-shot_Exemplar_Selection}. For SLM, we perform $3$ runs with $3$ different random seeds. 

%We run our each of our experiment three times and report the average and standard deviation of their domain-wise and overall Macro F1 on {\dataset}. 
%Given that, we do not fine-tune the LLMs, we report the same scores as the PLM baselines but from a single run. 
The results can be seen in Table \ref{table:baseline_results}. Our findings are described below.

\paragraph{Fine-tuning SLMs on combined training sets yields better {\dataset} performance} As we can see from the results, the SLMs fine-tuned on \textsc{SciNLI} and \textsc{MSciNLI} generally show a similar performances on the {\dataset} set. 
%for all PLMs, the performance fine-tuned on \textsc{SciNLI} and \textsc{MSciNLI} shows very similar domain-wise and overall performances on the {\dataset} test set. 
The performance shows consistent improvements across domains when the SLMs are fine-tuned on \textsc{MSciNLI+}, which is the combination of the training sets of \textsc{SciNLI} and \textsc{MSciNLI}. Therefore, fine-tuning the models using a training set with larger size and higher diversity enhances its robustness in an OOD setting. However, given that the best performing model with \textsc{SciBERT} shows a Macro F1 of only $78.17\%$, there is a substantial headroom for future improvements. 

%\vspace{-2mm}

\paragraph{Domain-specific pre-training is more useful for {\dataset} than `better' pre-training methods}
%\paragraph{Domain specific and `better' pre-training methods help in improving the performance on {\dataset}} 
The results show that \textsc{SciBERT}, \textsc{RoBERTa} and \textsc{XLNet} outperform \textsc{BERT} by a substantial margin in all domains. Note that the only difference between \textsc{BERT} and \textsc{SciBERT} is that \textsc{BERT} was pre-trained using generic text from Wikipedia and BookCorpus, whereas \textsc{SciBERT} was pre-trained using scientific text. Thus, the domain-specific pre-training of \textsc{SciBERT} aids in achieving a better performance than \textsc{BERT} on {\dataset}. Both \textsc{RoBERTa} and \textsc{XLNet} were pre-trained using general domain text similar to \textsc{BERT}. However, stronger (better) pre-training methods were employed in pre-training these two models and we observe their performance improvements for {\dataset} over \textsc{BERT}. 
% Comparing the performance among \textsc{SciBERT}, \textsc{RoBERTa}, and \textsc{XLNet}, 
We can also see that \textsc{XLNet} shows a substantially lower Macro F1 than \textsc{SciBERT} and \textsc{RoBERTa}. While the best performance results shown by \textsc{RoBERTa} and \textsc{SciBERT} (both when fine-tuned on \textsc{MSciNLI+}), are almost identical, \textsc{SciBERT} outperforms \textsc{RoBERTa} in several cases (e.g., when they are fine-tuned on \textsc{SciNLI} or \textsc{MSciNLI} separately). Thus, domain-specific pre-training (on scientific documents) results in a better performance for {\dataset} than `better' (more robust) pre-training methods.   

%Comparing the performance of \textsc{SciBERT} with \textsc{RoBERTa} and \textsc{XLNet}, we can see that \textsc{SciBERT} consistently outperforms \textsc{XLNet}, and . %This indicates that domain specific pre-training of \textsc{SciBERT} is more useful for {\dataset} than the stronger methods employed for pre-training \textsc{RoBERTa} and \textsc{XLNet}.  

\paragraph{Fine-tuned Small Language Models outperform Prompt-based Large Language Models} We can observe from Table \ref{table:baseline_results} that the SLMs perform much better than even the leading prompt-based LLMs (such as \textsc{GEMINI-1.5-PRO} and \textsc{GPT-4O}) on all three domains. The average performance gap is approximately 15\% between the highest performing SLM (\textsc{SCIBERT}) and the top-performing LLM. \textsc{GEMINI-1.5-PRO} and \textsc{GPT-4O} outperform open-source models in all few-shot settings, with \textsc{GEMINI-1.5-PRO} achieving the strongest overall performance in zero-shot settings, surpassing both \textsc{GPT-4O} and all open-source models.
Among open-source LLMs (Table \ref{table:baseline_results}), \textsc{PHI-3} demonstrates the best performance, outperforming \textsc{Llama-2}, \textsc{Llama-3}, and \textsc{Mistral} in both zero-shot and few-shot settings, indicating strong complex reasoning capabilities. Notably, \textsc{PHI-3} few-shot with \textsc{MSciNLI} exemplars shows the best performance among open-source models. While \textsc{GPT-4o}'s zero-shot capability was below \textsc{PHI-3}, it still outperformed other open-source baselines. The superiority of proprietary models is particularly evident in few-shot settings, where both \textsc{GEMINI-1.5-PRO} and \textsc{GPT-4o} show similar high performance and significantly outperform all open-source models, suggesting superior in-context learning ability for scientific NLI tasks. We show results with fine-tuned Llama-2 in Appendix \ref{appendix:Fine-Tuned_LLM_Results}.

\subsection{Analysis}

\paragraph{In-Domain vs. Out-of-Domain}
Given that we establish our baselines using the training sets of \textsc{SciNLI} and \textsc{MSciNLI}, i.e., sentence pairs from CS domains, the baseline performance results reported on {\dataset} in Table \ref{table:baseline_results} are in the out-of-domain (OOD) setting. We now compare the OOD performances with the in-domain (ID) performance of the respective models. The ID performance is calculated by evaluating the model on the test set of the dataset it is trained on. We choose both \textsc{SciBERT} and \textsc{RoBERTa} because of their strong performance on {\dataset}. The results can be seen in Table \ref{table:id_vs_ood}.
%We choose both \textsc{SciBERT} and \textsc{RoBERTa} for th We choose the \textsc{RoBERTa} model for this analysis as it is our best performing baseline. 

\setlength\dashlinedash{0.2pt}
\setlength\dashlinegap{1.5pt}
\setlength\arrayrulewidth{0.3pt}
\begin{table}[t]
\centering
\small

\scalebox{0.9}{
  \begin{tabular}{l c c}
    \toprule
    {\bf Model} & {\bf \textsc{ID}} & {\bf OOD ({\dataset})} \\
   \toprule

    \textsc{SciBERT}$_{SciNLI}$ & $77.11$ & $76.27$\\
    \textsc{SciBERT}$_{MSciNLI}$  & $76.66$ & $77.66$\\
    \textsc{SciBERT}$_{MSciNLI+}$ & $77.38$ & $78.17$ \\
    \hdashline
    \textsc{RoBERTa}$_{SciNLI}$ & $78.24$ & $75.43$\\
    \textsc{RoBERTa}$_{MSciNLI}$  & $77.02$ & $75.38$\\
    \textsc{RoBERTa}$_{MSciNLI+}$ & $78.77$ & $78.11$ \\
    \bottomrule
  \end{tabular}}
  %\vspace{-3mm}
  \caption{\small Macro F1 ($\%$) shown by \textsc{SciBERT} and \textsc{RoBERTa} in ID and OOD ({\dataset}) settings.}
  \vspace{-3mm}
  % \caption{Cross dataset performance of \textsc{BiLSTM} and \textsc{RoBERTa} models.}
    
    \label{table:id_vs_ood}

\end{table}

\setlength\dashlinedash{0.2pt}
\setlength\dashlinegap{1.5pt}
\setlength\arrayrulewidth{0.3pt}
\begin{table}[t]
\centering
\small

\scalebox{1.00}{
  \begin{tabular}{r c c c c}
    \toprule
    {\bf \#Shot} & {\bf \textsc{Psy}} & {\bf \textsc{Engg}} & {\bf \textsc{PH}} & {\bf \textsc{Overall}}\\
   \toprule

    $4$-\textsc{shot} & $58.64$ & $56.76$ & $55.68$ & $57.16$\\
    $8$-\textsc{shot}  & $58.80$ & $57.15$ & $56.82$ & $57.71$\\
    $12$-\textsc{shot}  & $59.63$ & $58.29$ & $56.69$ & $58.32$\\
    $16$-\textsc{shot}  & $59.56$ & $57.97$ & $56.50$ & $58.14$\\
    \bottomrule
  \end{tabular}}
  %\vspace{-3mm}
  \caption{\small $4$-shot, $8$-shot, $12$-shot  and $16$-shot Macro F1 (\%) by \textsc{Phi-3}. Here, \textsc{Psy}: \textsc{Psychology}, \textsc{Engg}: \textsc{Engineering}, and \textsc{PH}: \textsc{Public Health}.}
  \vspace{-3mm}
  % \caption{Cross dataset performance of \textsc{BiLSTM} and \textsc{RoBERTa} models.}
    
    \label{table:4_vs_8_shot}

\end{table}

\begin{table*}[t]
\centering
\small
% \addlinespace

\scalebox{0.8}{
  \begin{tabular}{l c c c c c}
    \toprule
    {\bf \diagbox{Model}{Dataset}} & {\bf \textsc{Sentence Input}} & {\bf {PSYCHOLOGY}} & {\bf \textsc{ENGINEERING}} & {\bf \textsc{PUBLIC HEALTH }} & {\bf \textsc{Macro Ave.}}\\
   \toprule
    \textsc{ROBERTA$_{MSciNLI+}$} & \textsc{BOTH SENTENCES} & $77.91$ & $77.63$ & $78.79$ & $78.11$\\
    \textsc{ROBERTA$_{MSciNLI+}$} & \textsc{ONLY 2nd SENTENCE} & $53.17$ & $58.59$ & $52.05$ & $54.64$\\
    \hdashline
    \textsc{SciBERT$_{MSciNLI+}$} & \textsc{BOTH SENTENCES} &  $79.18$ & $76.50$ & $78.79$ & $78.17$\\
    \textsc{SciBERT$_{MSciNLI+}$} & \textsc{ONLY 2nd SENTENCE} &  $56.68$ & $58.12$ & $54.54$ & $56.50$\\
    \bottomrule
    
  \end{tabular}}

   \caption{\small Comparison of Macro F1 scores (\%) for RoBERTa and SciBERT on MISMATCHED domains when using both premise and hypothesis sentences versus only the hypothesis sentence as input.
  }
  
    \label{table:Hypothesis-only_baseline}
\end{table*}

First, we can observe that RoBERTa which is trained in a robust way shows a consistent drop in performance from ID to OOD especially when the model is fine-tuned individually on SciNLI or MSciNLI showing about 2-3\% performance drop. These results demonstrate that the OOD is more challenging for RoBERTa. Second, training RoBERTa on increased diversity data (i.e., MSciNLI+) lowers the gap in performance between ID and OOD. These results show the impact of data diversity on model training and generalization to OOD data. Third, we can observe that while the SciBERT model (which is trained on scientific documents) performs worse than RoBERTa on ID data, its scientific knowledge that is learned from large training sets of research papers during its pre-training is beneficial for the scientific OOD data and in fact, it helps the SciBERT model to achieve similar performance with that of RoBERTa. These results show that scientific knowledge that is learned during the pre-training of SciBERT is retained and leveraged in the OOD setting and is more beneficial in OOD than training the model in a more robust way but on general (not specifically scientific) data. Similar to RoBERTa, we observe that the results with SciBERT in OOD when trained with increased diversity data (i.e., MSciNLI+) show that this diversity is beneficial on the OOD data.

% As we can see, the ID performance of \textsc{RoBERTa} is consistently better than its OOD performance on {\dataset}. That is, regardless of the level of diversity in training and testing data, \textsc{RoBERTa} shows a better performance in the ID setting compared to OOD. On the other hand, when \textsc{SciBERT} has access to diverse training data (\textsc{MSciNLI+}), it yields higher performance compared with the settings when it is trained with only SciNLI or MSciNLI separately. Interestingly, although it shows a slightly worse ID performance than RoBERTa, it is able to achieve a similar OOD performance as \textsc{RoBERTa}. %However, unlike \textsc{RoBERTa}, \textsc{SciBERT} shows a lower ID performance when it is trained and tested on diverse data (\textsc{MSciNLI} and \textsc{MSciNLI+}) compared to its OOD performance on {\dataset}. 

%The results can be seen in Table \ref{table:id_vs_ood}. We find that:

%\paragraph{ID performance is consistently better than OOD performance.} The results show that the ID performance of all three models (fine-tuned on \textsc{SciNLI}, \textsc{MSciNLI} and \textsc{MSciNLI+}) is higher than their respective OOD performance on {\dataset}. Therefore, the domains in our dataset exhibits unique linguistic characteristics which are not captured by the existing scientific NLI training sets resulting in a drop in performance. As a result, {\dataset} can indeed serve as a challenging OOD test-bed for scientific NLI. 

\begin{table*}[t]
\centering
\small
% \addlinespace

\scalebox{0.9}{
  \begin{tabular}{l c c c c c}
    \toprule
%      &  \multicolumn{2}{c}{\bf SICK} & {\bf SciTail} & SNLI & SciNLI\\%\cline{2-4}
{\bf \textsc{Model}} &  {\bf \textsc{Contrasting}} &  {\bf \textsc{Reasoning}} & {\bf \textsc{Entailment}} & {\bf \textsc{Neutral}} & {\bf \textsc{Macro Ave.}}\\
%   {\bf Model} & {\bf F1}   & {\bf F1} & \textsc{\bf F1} &
%   {\bf F1} & {\bf Macro F1} & \textsc{\bf Acc}\\ %\cline{2-4}
    \midrule
    \textbf{SciBERT}\\
    \mbox{{\color{white}{x}}}\textsc{Psychology} & $81.60 \pm 0.9$ & $74.15 \pm 1.1$ & $79.97 \pm 1.5$ & $80.99 \pm 0.2$ & $79.18 \pm 0.4$\\
    \mbox{{\color{white}{x}}}\textsc{Engineering}  & $80.98 \pm 0.4$ & $76.50 \pm 1.1$ & $75.09 \pm 1.4$ & $73.43 \pm 1.0$ & $76.50 \pm 0.8$\\
    \mbox{{\color{white}{x}}}\textsc{Public Health} & $80.25 \pm 0.4$ & $74.55 \pm 0.3$ & $80.09 \pm 1.0$ & $80.31 \pm 0.9$ & $78.79 \pm 0.3$\\
    \mbox{{\color{white}{x}}}\textsc{\dataset} & $80.94 \pm 0.5$ & $75.09 \pm 0.6$ & $78.44 \pm 0.3$ & $78.22 \pm 0.6$ & $78.17 \pm 0.2$\\
    \hdashline
    \textbf{Phi-3}\\
    \mbox{{\color{white}{x}}}\textsc{Psychology} & $70.28 \pm 1.37$ & $40.60 \pm 2.95$ & $62.44 \pm 0.62$ & $61.27 \pm 0.94$ & $58.65 \pm 1.10$\\
    \mbox{{\color{white}{x}}}\textsc{Engineering} & $71.35 \pm 1.30$ & $47.10 \pm 2.90$ & $51.20 \pm 2.07$ & $57.42 \pm 0.31$ & $56.77 \pm 0.57$\\
    \mbox{{\color{white}{x}}}\textsc{Public Health} & $67.53 \pm 0.73$ & $44.08 \pm 4.09$ & $52.51 \pm 2.27$ & $58.62 \pm 2.11$ & $55.68 \pm 0.25$\\
    \mbox{{\color{white}{x}}}\textsc{\dataset} & $69.67 \pm 0.40$ & $44.07 \pm 3.27$ & $55.92 \pm 0.89$ & $59.00 \pm 0.97$ & $57.16 \pm 0.59$\\
    \bottomrule
    
  \end{tabular}}

   \caption{\small Class-wise F1 (\%) and their macro averages (\%) of our best performing SLM and LLM baselines on each domain in {\dataset} and their combination.
  }
  
    \label{table:class_wise_results}
\vspace{-3mm}
\end{table*}

\paragraph{Few-shot Scaling Experiments}
While proprietary models (\textsc{GEMINI-1.5-PRO} and \textsc{GPT-4O}) demonstrated superior performance, we selected \textsc{PHI-3} for few-shot scaling experiments as the best performing open-source model. This choice enables comprehensive analysis of in-context learning mechanisms with full reproducibility and extensive experimentation without API constraints. We used exemplars from \textsc{MSciNLI} given its superior performance on {\dataset} among all settings (as discussed in \ref{sec:results}). Our experiments investigate the impact of increasing few-shots from 4 to 8, 12, and 16 on \textsc{PHI-3's} performance, with results shown in Table \ref{table:4_vs_8_shot}. Results show that $12$-shots achieve slightly better performance than $4$-shots and $8$-shots, while performance drops at $16$-shots. %Thus, increasing the number of shots in the prompt slightly improves the performance of \textsc{Phi-3} on {\dataset}}.

\paragraph{Hypothesis-only Baseline Experiment} To verify whether our dataset contains spurious correlations or not, i.e., any stylistic artifacts that are present only in the hypotheses and are indicative of the label (without the need for the premise), we compare hypothesis-only models against full premise-hypothesis models using \textsc{RoBERTa} and \textsc{SciBERT} fine-tuned on \textsc{MSciNLI+} in Table \ref{table:Hypothesis-only_baseline}. We chose \textsc{MSciNLI+} as the training set because fine-tuned models (\textsc{RoBERTa} and \textsc{SciBERT}) achieved their highest performance on \dataset when trained on \textsc{MSciNLI+} compared to \textsc{SciNLI} or \textsc{MSciNLI} alone (Table \ref{table:baseline_results}). Results show significant performance degradation when using only the hypothesis compared to the full premise-hypothesis input, demonstrating that premise-hypothesis understanding is critical for model performance. Thus, our dataset does not exhibit hypothesis-only artifacts.

\paragraph{Class-wise Performance} We report the per-class F1 scores of our best performing SLM baseline, \textsc{SciBERT} (fine-tuned using \textsc{MSciNLI+}) and best performing LLM baseline, \textsc{Phi-3} (in the few-shot setting with \textsc{MSciNLI} exemplars) in Table \ref{table:class_wise_results}. We can see that generally, both types of models show lower F1 scores for the \textsc{Reasoning} class compared with the other classes. Therefore, recognizing a \textsc{Reasoning} relation between sentences is more challenging than recognizing other scientific NLI relations. We provide an in-depth analysis of the ``reasoning'' relation in Appendix \ref{reasoning_analysis}.

\section{Harnessing Implicit Relations}
\label{section:implicit_relations}
%Given that all baselines for {\dataset} utilize out-of-domain (OOD) data due to a lack of in-domain (ID) training examples, we aim to harness the ID sentence pairs which do not contain any linking phrases between them but implicitly establishes a relation. Our method for harnessing the implicit relations is described below:
Existing training sets for scientific NLI datasets (i.e., \textsc{SciNLI} and \textsc{MSciNLI}) only include sentence pairs where the relation between them is made \textit{explicit} with linking phrases. We posit that, if two sentences are adjacent, there can potentially be a scientific NLI relation between them despite the second sentence not starting with a linking phrase. We define the relation between these sentence pairs as an \textit{implicit} relation. Here, we propose to incorporate adjacent sentences with \textit{implicit} relations in model training and analyze their impact on models' performance. We detail below the data sources from which we extract implicit sentence pairs, how we annotate them, and how we use them in model training. 

\paragraph{Data} The implicit sentence pairs are sourced from the research papers from \textsc{SciNLI}, \textsc{MSciNLI} and {\dataset} separately. For each dataset, we extract the adjacent sentence pairs in which none of the sentences contain any linking phrases as the examples potentially containing an \textit{implicit} \textsc{Entailment}/\textsc{Contrasting}/\textsc{Reasoning} relation. For the \textsc{Neutral} class, we randomly pair two non-adjacent sentences selected from the other three classes. For \textsc{SciNLI} and \textsc{MSciNLI}, we extracted the implicit pairs from papers that are part of the training set, whereas for {\dataset}, we extracted the implicit pairs from papers that are not utilized to construct its test and development sets. We extracted $\approx 210K$ and $\approx 120K$ implicit sentence pairs for \textsc{SciNLI}/\textsc{MSciNLI} and \textsc{\dataset} respectively, with the number of implicit relations being about twice as large as explicit relations. We provide examples of implicit sentence pairs extracted from different domains in our {\dataset} dataset in Appendix \ref{implicit}.

\setlength\dashlinedash{0.2pt}
\setlength\dashlinegap{1.5pt}
\setlength\arrayrulewidth{0.3pt}
\begin{table}[t]
\centering
\small

\scalebox{0.9}{
  \begin{tabular}{l c c}
    \toprule
    {\bf \diagbox{Model}{Dataset}} & {\bf \textsc{SciNLI}} & {\bf {\dataset}} \\
   \toprule

    \textsc{SciBERT$_{MS+}$} & $79.04$ & $78.17$\\
    \textsc{SciBERT$_{MS+}$ $_{+}$ $_{Impl}$} & ${\bf 79.44}$ & ${\bf 79.66}$\\
    \hdashline
     %\textsc{BiLSTM} & \textsc{MSciNLI} & $54.72$ & $54.40$\\
    % \textsc{Phi-3}$_{fs-S}$ & $-$ & $53.71$ \\
    % \textsc{Phi-3}$_{fs-MS}$ & $-$ & $56.19$ \\
    \textsc{Phi-3}$_{fs-Expl\{SciNLI\}}$ & $59.67 \pm 1.92$ & $55.84 \pm 0.98$ \\
    \textsc{Phi-3}$_{fs-Impl\{SciNLI\}}$ & ${\bf 61.41} \pm 1.11$ & ${\bf 56.56} \pm 1.47$ \\
    \hdashline
    \textsc{Phi-3}$_{fs-Expl\{MSciNLI\}}$ & $59.88 \pm 1.15$ & $57.16 \pm 0.59$ \\
   \textsc{Phi-3}$_{fs-Impl\{MSciNLI\}}$ & ${\bf 60.58} \pm 0.43$ & ${ \bf 57.50} \pm 0.06$ \\
    \hdashline
   %  \textsc{Phi-3}$_{fs-Expl\{MS+\}}$ & $61.06 \pm 1.00$ & $56.87 \pm 0.25$ \\
   % \textsc{Phi-3}$_{fs-Impl\{MS+\}}$ & ${\bf 61.70 \pm 0.48}$ & ${ \bf 57.42 \pm 0.73}$ \\
   %  \hdashline
    \textsc{Phi-3}$_{fs-Expl\{MisMatched\}}$ & $58.57 \pm 1.29$ & $56.96 \pm 0.68$ \\
    \textsc{Phi-3}$_{fs-Impl\{MisMatched\}}$ & ${\bf 61.03 \pm 0.40}$ & ${\bf 58.26 \pm 0.25}$ \\
    \bottomrule
  \end{tabular}}
  %\vspace{-3mm}
  \caption{\small Performance comparison between models utilizing \textit{implicit} relations with models only using \textit{explicit} examples. Here, MS+: \textsc{MSciNLI+}, Expl: \textit{explicit} and Impl: \textit{implicit}.}
  %\vspace{-5mm}
  % \caption{Cross dataset performance of \textsc{BiLSTM} and \textsc{RoBERTa} models.}
    
    \label{table:implcit_results}

\end{table}

\paragraph{Implicit Relation Annotation} Next, we identify the implicit scientific NLI relation among the extracted sentence pairs in three steps: a) assign pseudo-labels to the extracted sentence pairs based on the predictions made by the \textsc{SciBERT} model fine-tuned on \textsc{MSciNLI+};
% {\color{blue} \textsc{ChatGPT-4}, a \textsc{Large Language Model (LLM)}}; 
b) filter the examples based on a confidence (i.e., the probability for the predicted pseudo-label by the model) threshold of $0.6$; and c) filter the examples where a \textsc{Contrasting/Entailment/Reasoning} label is predicted for a non-adjacent sentence pair or a \textsc{Neutral} label for an adjacent sentence pair.  

\paragraph{Incorporating Implicit Relations} We %evaluate the impact of 
incorporate \textit{implicit} relations in model training by experimenting with \textsc{SciBERT} and \textsc{Phi-3} and evaluating their performance on the test sets of \textsc{SciNLI} and \textsc{\dataset}. For \textsc{SciBERT}, we first fine-tune an \textit{out-of-the-box} model using the selected \textit{implicit} examples from the same domain as the test set (i.e., when the test set is \textsc{\dataset}, the implicit examples are from papers from the \textsc{\dataset} domains). We then continue fine-tuning the model using the \textit{explicit} examples from \textsc{MSciNLI+}. For \textsc{Phi-3}, we randomly sample four examples (one from each class) from the selected \textit{implicit} set, and use them as the exemplars in the few-shot setting.  

% \setlength\dashlinedash{0.2pt}
% \setlength\dashlinegap{1.5pt}
% \setlength\arrayrulewidth{0.3pt}
% \begin{table}[t]
% \centering
% \small

% \scalebox{1.00}{
%   \begin{tabular}{l c c}
%     \toprule
%     {\bf \diagbox{Model}{Dataset}} & {\bf \textsc{SciNLI}} & {\bf {\dataset}} \\
%    \toprule

%     \textsc{RoBERTa$_{MS+}$} & $79.51$ & $77.37$\\
%     \textsc{RoBERTa$_{MS+}$ $_{+}$ $_{Impl}$} & $79.40$ & $77.52$\\
%     \hdashline
%      %\textsc{BiLSTM} & \textsc{MSciNLI} & $54.72$ & $54.40$\\
%     \textsc{Llama-2}$_{fs-S}$ & $49.77$ & $46.77$ \\
%     \textsc{Llama-2}$_{fs-MS}$ & $51.19$ & $47.55$ \\
%     \textsc{Llama-2}$_{fs-Impl}$ & $45.81$ & $45.18$ \\
%     \bottomrule
%   \end{tabular}}
%   %\vspace{-3mm}
%   \caption{\small Performance comparison between models utilizing \textit{implicit} relations with models only using \textit{explicit} examples. Here, MS+: \textsc{MSciNLI+}, S: \textsc{SciNLI}, MS: \textsc{MSciNLI}, and Impl: \textit{implicit}.}
%   \vspace{-5mm}
%   % \caption{Cross dataset performance of \textsc{BiLSTM} and \textsc{RoBERTa} models.}
    
%     \label{table:implcit_results}

% \end{table}

\paragraph{Results} Table \ref{table:implcit_results} shows a comparison between the models that use only \textit{explicit} examples 
% from {\color{blue}\textsc{SciNLI}, \textsc{MSciNLI}, \textsc{MSciNLI+}, \textsc{\dataset} and the models 
with their counterparts that incorporate % \textit{explicit} and/or 
\textit{implicit} examples. As we can see, the Macro F1 of \textsc{SciBERT} improves by $1.5\%$ for {\dataset} when \textsc{implicit} relations are incorporated into model training. In addition, the performance of \textsc{Phi-3} also shows improvement in Macro F1 when \textit{implicit} examples are used as the few-shot exemplars compared to \textit{explicit} examples from \textsc{SciNLI}, \textsc{MSciNLI}, %\textsc{MSciNLI+}, 
\textsc{\dataset} used as exemplars. Given that all sentence pairs from \textsc{SciNLI}, \textsc{MSciNLI} and \textsc{MSciNLI+} are out-of-domain for {\dataset}, incorporating in-domain \textit{implicit} relations into models' training helps improve its performance. Interestingly, when \textsc{Phi-3} with few-shot exemplars from {\dataset} is evaluated on \textsc{SciNLI}, we can see an improvement of $2.46$ (from $58.57$ to $61.03$) which demonstrates the benefits of using implicit relations that make the model more robust and capable to generalize better.
% For \textsc{SciNLI}, we see a moderate improvement by both \textsc{SciBERT} and \textsc{Phi-3} when \textit{implicit} examples are utilized. %Moreover, both \textit{explicit} and \textit{implicit} exemplars show a similar few-shot performance on \textsc{SciNLI}. 
% Recall that, \textsc{MSciNLI+} already contains a large number of \textit{explicit} relations from the same domain as \textsc{SciNLI} (i.e., ACL). {\color{blue}However, incorporating \textit{implicit} relations results in a degree of improvement that is greater than that observed with} {\dataset}. 
Thus, given the improvements for both datasets, 
we can conclude that sentence pairs with \textit{implicit} relations can be a valuable resource for exposing scientific NLI models to more diverse data that can further improve the performance.

%We can see a comparison of performance (Macro F1) between the \textsc{SciBERT} model using \textit{explicit} examples from \textsc{MSciNLI+} with the \textsc{SciBERT} model using \textit{implicit} examples (in addition to \textit{explicit} examples for \textsc{RoBERTa}) in Table \ref{table:implcit_results}.

%\paragraph{Observations} \mobashir{needs to be updated} We can see a comparison of performance (Macro F1 and Accuracy) between the models using \textit{explicit} examples with the models using \textit{implicit} examples (in addition to \textit{explicit} examples for \textsc{RoBERTa}) in Table \ref{table:implcit_results}. The results show that for both \textsc{SciNLI} and {\dataset}, the \textsc{RoBERTa} model shows a very similar performance with and without incorporating the \textit{implicit} relations. That is, the usage of additional examples containing \textit{implicit} relations fail to show any improvement. Furthermore, for \textsc{Llama-2}, we can see that the few-shot exemplars from the selected \textit{implicit} set shows a lower performance than the \textit{explicit} exemplars from \textsc{SciNLI} and \textsc{MSciNLI} training sets. Therefore, the signal from the linking phrases is strong and highly important for improving the performance in scientific NLI.

%\subsection{Class-wise Performance}
\vspace{-1mm}
\section{Conclusion \& Future Directions}
In this paper, we introduce a {\dataset} test-bed for scientific NLI, derived from non-CS domains unlike the existing datasets. We establish strong baselines on the {\dataset} set with both SLMs and LLMs using the training sets from \textsc{SciNLI} and \textsc{MSciNLI}. Our results show that the best performing baseline achieves a Macro F1 of only $78.17\%$, illustrating the substantial room for future improvements. Furthermore, we show that sentence pairs containing an \textit{implicit} scientific NLI relation can aid in improving the performance of two scientific NLI benchmarks. In our future work, we will develop domain adaptation methods for scientific NLI to improve the performance on the {\dataset} set.

\section*{Acknowledgments} 
We thank US-NSF for support from grant IIS-2107518 and UIC Discovery Partners Institute which supported the research and the computation in this study. %We also thank our reviewers for their insightful feedback and comments.
 Research reported in this publication was also partially supported by the CNAP Center of Biomedical Research Excellence of the NIH under grant No. P20GM113109. %The content is solely the responsibility of the authors and does not necessarily represent the official views of the NIH.

%we provide a thorough analysis of incorporating \textit{implicit} relations in model training and find that they can improve the performance of scientific NLI models improvements over only using \textit{explicit} examples. 

\section*{Limitations}
Our {\dataset} benchmark indeed enhances the diversity in scientific NLI to non-CS domains. However, there are numerous scientific domains and disciplines (e.g., Physics, Chemistry, etc.) that are not covered by our dataset. Therefore, a future research direction is to study scientific NLI to other non-CS domains that can serve as a more robust and generalized benchmark.

%further extending scientific NLI to other domains can be a future direction

\bibliography{anthology,custom}
\bibliographystyle{acl_natbib}

%We have also explored an experiment where we first fine-tune the model using \textit{MSciNLI+} and then continue fine-tun

% \paragraph{Model Fine-tuning} After selecting the sentence pairs with implicit relations, we fine-tune the pre-training PLMs and LLMs (i.e., \textsc{RoBERTa} and \textsc{Llama-2}). We then continue the fine-tuning of these models using the training sets with explicit relations (i.e., \textsc{MSciNLI+ (S)}). 

% \section{Experiments \& Results}

% \subsection{Baselines}
% For establishing the baselines using OOD sentence pairs with explicit relations, we fine-tune both \textsc{RoBERTa} and \textsc{Llama-2} using  a) \textsc{SciNLI}; b) \textsc{MSciNLI}; c) a combination of \textsc{SciNLI} and \textsc{MSciNLI}, denoted as \textsc{MSciNLI+}; d) and a combination of \textsc{ACL - small} and \textsc{MSciNLI}, denoted as \textsc{MSciNLI+ (s)}. Here, \textsc{ACL - small} is a downsampled version of the \textsc{SciNLI} training set to a size equal to the other domains in \textsc{MSciNLI}. Our choice of these two models are based on the fact that \textsc{RoBERTa}, \textsc{Llama-2} have been reported as the best performing PLM and LLM baselines, respectively on recently published \textsc{MSciNLI}. We perform the experiments with \textsc{RoBERTa} 3 times with different random seeds and report their average and standard deviation. Given the high computation cost incurred while fine-tuning \textsc{Llama-2}, we report the results from a single run. The performance of our baselines can be seen in Table X. Our findings are described below:

\clearpage
\newpage
\newpage
\appendix

\begin{table*}
\centering
\resizebox{\textwidth}{!}{\begin{tabular}{l|l|l|l|l|l|r} 
\hline
\multicolumn{1}{c|}{\textbf{Dataset}} & \multicolumn{1}{c|}{\textbf{Source/Domains}} & \textbf{Classes} & \multicolumn{1}{c|}{\textbf{ID}} & \multicolumn{1}{c|}{\textbf{OOD}} & \textbf{Hypothesis} & \multicolumn{1}{c}{\textbf{$\approx$ Size}}  \\ 
\hline

\textsc{RTE} \cite{wang2018glue}  & Wikipedia and news sources & 2 $|$  \begin{tabular}[c]{@{}l@{}}{\it entailment},\\ {\it non-entailment} \end{tabular}  & {\color{blue}\Checkmark} & {\color{orange}\xmark} & Synthetic & 2,500 \\
  \hdashline
\textsc{SICK} \cite{marelli-etal-2014-sick} & Image captions and video descriptions & 3 $|$ \begin{tabular}[c]{@{}l@{}}{\it entailment}, \\{\it contradiction}\\ {\it neutral} \end{tabular} & {\color{blue}\Checkmark} & {\color{orange}\xmark} & Synthetic  & 10,000 \\
  \hdashline
\textsc{SNLI} \cite{bowman-etal-2015-large} & Image captions & 3 $|$ \begin{tabular}[c]{@{}l@{}}{\it entailment}, \\{\it contradiction}\\ {\it neutral} \end{tabular}  & {\color{blue}\Checkmark} & {\color{orange}\xmark} & Synthetic & 570,000 \\
 \hdashline
\textsc{MultiNLI}  \cite{williams-etal-2018-broad}  & \begin{tabular}[c]{@{}l@{}}Nine sources from second  OANC release \\(Face-to-face, government, letter, etc.) \\ \& Fiction (mystery, humor, western, etc.)\end{tabular} & 3 $|$ \begin{tabular}[c]{@{}l@{}}{\it entailment},\\ {\it contradiction}\\ {\it neutral} \end{tabular}  & {\color{blue}\Checkmark} & {\color{blue}\Checkmark} & Synthetic & 433,000 \\
 \hdashline
\textsc{ANLI} \cite{nie-etal-2020-adversarial}  & Wikipedia, news, fiction,  spoken text, etc. & 3 $|$  \begin{tabular}[c]{@{}l@{}}{\it entailment},\\ {\it contradiction}\\ {\it neutral} \end{tabular}  & {\color{blue}\Checkmark} & {\color{orange}\xmark} & Synthetic & 170,000 \\
\hline
\textsc{MedNLI} \cite{romanov-shivade-2018-lessons}  & \begin{tabular}[c]{@{}l@{}}MIMIC-III, clinical notes \\(Past Medical History) \end{tabular} & 3 $|$  \begin{tabular}[c]{@{}l@{}}{\it entailment}\\ {\it contradiction}\\ {\it neutral} \end{tabular}  & {\color{blue}\Checkmark} & {\color{orange}\xmark} & Real & 14,000 \\
 \hdashline
\textsc{NLI4CT} dataset \cite{jullien2023nli4ct}  & \begin{tabular}[c]{@{}l@{}}Breast cancer clinical trial  reports \\ (U.S. National Library of Medicine) \end{tabular} & 2 $|$  \begin{tabular}[c]{@{}l@{}}{\it entailment}\\ {\it contradiction} \end{tabular}  & {\color{blue}\Checkmark} & {\color{orange}\xmark} & Synthetic & 2,400 \\
 \hdashline
\textsc{NLI4CT-P} \cite{jullien2024semeval}  & \begin{tabular}[c]{@{}l@{}}Breast cancer clinical trial  reports \\ (U.S. National Library of Medicine) \end{tabular}  & 2 $|$  \begin{tabular}[c]{@{}l@{}}{\it entailment}\\ {\it contradiction} \end{tabular}  & {\color{blue}\Checkmark} & {\color{orange}\xmark} & Synthetic & \begin{tabular}[c]{@{}l@{}}8,600 \end{tabular} \\
\hline
\textsc{SciNLI} \cite{sadat-caragea-2022-scinli}  & \begin{tabular}[c]{@{}l@{}}Research articles from ACL Anthology \end{tabular}  & 4 $|$  \begin{tabular}[c]{@{}l@{}}{\it entailment} \\ {\it reasoning} \\ {\it contrasting} \\ {\it neutral} \end{tabular}  & {\color{blue}\Checkmark} & {\color{orange}\xmark} & Real & \begin{tabular}[c]{@{}l@{}}101,000\end{tabular} \\
 \hdashline
\textsc{MSciNLI} \cite{sadat-caragea-2024-mscinli}  & \begin{tabular}[c]{@{}l@{}}Computer science research articles, \\ \textsc{Hardware}, \textsc{Networks},\\ \textsc{Software} \& \textsc{its engineering}, etc.\end{tabular}  & 4 $|$  \begin{tabular}[c]{@{}l@{}}{\it entailment} \\ {\it reasoning} \\ {\it contrasting} \\ {\it neutral} \end{tabular}  & {\color{blue}\Checkmark} & {\color{orange}\xmark} & Real & \begin{tabular}[c]{@{}l@{}}127,000\end{tabular} \\
 \hdashline
{\bf {\dataset}  (ours)}  & \begin{tabular}[c]{@{}l@{}}Research articles from \textsc{Public health} \\ \textsc{Psychology} and \textsc{Engineering}\end{tabular}  & 4 $|$  \begin{tabular}[c]{@{}l@{}}{\it entailment} \\ {\it reasoning} \\ {\it contrasting} \\ {\it neutral} \end{tabular}  & {\color{orange}\xmark}  & {\color{blue}\Checkmark}  & Real & \begin{tabular}[c]{@{}l@{}}2,700\end{tabular} \\
\hline
% \textsc{SciFact} \cite{wadden2020fact}  & \begin{tabular}[c]{@{}l@{}}Scientific articles (topics range from \\molecular biology to public health) \end{tabular} & 3 $|$  \begin{tabular}[c]{@{}l@{}}{\it supports} \\ {\it refutes} \\ {\it NEI}  \end{tabular}  & {\color{blue}\Checkmark}  & {\color{orange}\xmark}  & Synthetic & \begin{tabular}[c]{@{}l@{}}1,400\end{tabular} \\
%  \hdashline
% \textsc{HealthVer} \cite{sarrouti-etal-2021-evidence-based}  & \begin{tabular}[c]{@{}l@{}}Snippets returned by a search engine \\ and research articles \end{tabular} & 3 $|$  \begin{tabular}[c]{@{}l@{}}{\it supports} \\ {\it refutes} \\ {\it neutral}  \end{tabular}  & {\color{blue}\Checkmark}  & {\color{orange}\xmark}  & Real & \begin{tabular}[c]{@{}l@{}}14,000\end{tabular} \\ 
%  \hdashline
% \textsc{COVIDFact} \cite{saakyan2021covid}  & \begin{tabular}[c]{@{}l@{}}  Subreddit r/COVID19 and corresponding \\ links to scientific articles (COVID-19) \end{tabular} & 2 $|$  \begin{tabular}[c]{@{}l@{}}{\it supported} \\ {\it refuted}  \end{tabular}  & {\color{blue}\Checkmark}  & {\color{orange}\xmark}  & Both & \begin{tabular}[c]{@{}l@{}}4,000\end{tabular} \\ 
%  \hdashline
% \textsc{SciFact-Open} \cite{saakyan2021covid}  & \begin{tabular}[c]{@{}l@{}}  Scientific claims from \textsc{SciFact} and research \\abstracts from S2ORC, arXiv, or PubMed  \end{tabular} & 2 $|$  \begin{tabular}[c]{@{}l@{}}{\it supported} \\ {\it refuted}  \end{tabular}  & {\color{orange}\xmark}  & {\color{blue}\Checkmark}  & Real & \begin{tabular}[c]{@{}l@{}}500,000\end{tabular} \\ 
% \hline
\end{tabular}}
\caption{Comparison of relevant NLI datasets. The {\it Source/Domains} column indicates the sources of data collection and/or the domains covered by the dataset. The {\it Classes} column indicates the number of classes, followed by specific classes in the dataset. The {\it ID} and {\it OOD} columns indicate if the dataset is {\it in-domain} (i.e., contains both training and test data for some domains) and/or {\it out-of-domain} (i.e., contains only test data for some domains. {\it Hypothesis} refers to the fact that the hypothesis is {\it Real} (extracted directly from existing text) or {\it Synthetic} (written or re-written by human annotators). Finally, the last column, $\approx$ {\it Size} refers to the approximate numbers of pairs in  the dataset (note that some datasets may have a smaller number of premises).}
\label{tab:related_works}
\end{table*}

\section{Datasets for NLI}
\label{appendix-rel-work}
Table \ref{tab:related_works} shows a comparison of relevant datasets in terms of sources from which data was collected, domains covered,  classes, in-domain (ID) and out-of-domain (OOD) training, real or synthetic (generated) hypothesis and dataset size (as number of sentence pairs).

\section{Details on Data Annotation}

\subsection{Linking Phrases Used in Distant Supervision}
\label{appendix:linking_phrases}

\begin{table}
\centering
\small

\begin{tabular}{lp{0.3\textwidth}}
\toprule
\textbf{Class} & \textbf{Linking Phrases}\\
\midrule
\textsc{Contrasting} & ‘However’, ‘On the other hand', ‘In contrast', ‘On the contrary'\\
 \midrule
\textsc{Reasoning} & ‘Therefore', ‘Thus', ‘Consequently’, ‘As a result', ‘As a consequence', ‘From here, we can infer’\\
\midrule
\textsc{Entailment} & ‘Specifically’, ‘Precisely’, ‘In particular’, ‘Particularly’, ‘That is’, ‘In other words’\\
\bottomrule
\end{tabular}
\caption{Linking phrases used to extract sentence pairs and their corresponding classes.}
\label{table:linking_phrases}
%\vspace{-1mm}
\end{table}

The linking phrases and their classes used in the distant supervision method for automatically extracting and annotating sentence pairs in {\dataset} can be seen in Table \ref{table:linking_phrases}.

\subsection{Details about Annotators and Inter-Annotator Agreement}
\label{appendix: annotator_details}
We hire separate annotators for each of the three domains in our dataset via a cloud-sourcing platform called COGITO\footnote{\url{https://www.cogitotech.com/}}. For each domain, we complete $3$ pilot batches containing $52$ sentence pairs (balanced over classes). After each pilot batch, we provide feedback to the annotators on their work and ask them for their acknowledgement of our feedback before starting the next batch. The annotators are paid at a rate of \$$0.6$/sample.

The inter-annotator agreement varied across domains, as shown in Table \ref{table:IAA_per_domain}. \textsc{Psychology} showed the highest agreement (\textsc{Fleiss-K} = $0.78$), followed by \textsc{Engineering} ($0.70$) and \textsc{Public Health} ($0.65$). The variation in agreement rates likely reflects the differing complexity and ambiguity levels inherent to scientific texts across these domains.

\setlength\dashlinedash{0.2pt}
\setlength\dashlinegap{1.5pt}
\setlength\arrayrulewidth{0.3pt}
\begin{table}[t]
\centering
\small

\scalebox{1.00}{
  \begin{tabular}{r c c c }
    \toprule
    {\bf Domain} & {\bf \textsc{Psy}} & {\bf \textsc{Eng}} & {\bf \textsc{PH}}\\
   \toprule

    \textsc{Fleiss-k} & $0.78$ & $0.70$ & $0.65$\\
    \bottomrule
  \end{tabular}}
  %\vspace{-3mm}
  \caption{Inter-annotator agreement (\textsc{Fleiss-k}) by domain. Here, \textsc{Psy}: \textsc{Psychology}, \textsc{Engg}: \textsc{Engineering}, and \textsc{PH}: \textsc{Public Health}.}
  \vspace{-3mm}
    
    \label{table:IAA_per_domain}

\end{table}

\setlength\dashlinedash{0.2pt}
\setlength\dashlinegap{1.5pt}
\setlength\arrayrulewidth{0.3pt}
\begin{table}[t]
\centering
\small

  \begin{tabular}{l c c }
    \toprule
{\bf Class} & {\bf \#Annotated}     & {\bf Agreement} \\ %\cline{2-4}
   \midrule
    \textsc{Contrasting} & $744$ & $93.5\%$ \\
    \textsc{Reasoning} & $744$ & $93.4\%$ \\
    \textsc{Entailment} & $744$ & $96.2\%$ \\
    \textsc{Neutral} & $1021$ & $68.3\%$ \\
    \hdashline
    Overall & $3253$ & $85.7\%$\\
    \bottomrule
  \end{tabular}

  \caption{\small Number of sentence pairs annotated manually for each class and their agreement rate between the gold labels and automatically assigned labels.
  }

    \label{table:class_wise_agreement_table}
    
\end{table}

\setlength\dashlinedash{0.2pt}
\setlength\dashlinegap{1.5pt}
\setlength\arrayrulewidth{0.3pt}
\begin{table*}[t]

\centering
\small

\scalebox{0.95}{
  \begin{tabular} { p{5em} p{17em}  p{17em} p{5em}}
    \toprule
%    \cline{1-2}
   \multicolumn{1}{l}{\rule{0pt}{1ex}\textbf{Domain}} & \multicolumn{1}{c}{\rule{0pt}{1ex}\textbf{First Sentence}} & \multicolumn{1}{c}{\textbf{Second Sentence}}&
   \multicolumn{1}{c}{\rule{0pt}{1ex}\textbf{Class}}\\
%   \cline{1-2}
    \hline%\rule{0pt}{3ex}
    \textsc{Engineering} & Tools to predict its vibratory and acoustic performance at the design stage need to be developed. & an improved finite element model has been developed to analyse the vibration behaviour of a Permanent Magnet Synchronous Machine of a lift installation using the finite element software ABAQUS. &  \textsc{Reasoning}\\
    \hdashline
    \textsc{Psychology}  & This literature review provides information for identifying children who have been abused and neglected but exposes the need for a comprehensive screening instrument or protocol that will capture all forms of child abuse and neglect. & screening needs to be succinct, user-friendly, and amenable for use with children at every point of care in the healthcare system. &  \textsc{Contrasting}\\

    \bottomrule
    
  \end{tabular}}
  %\vspace{-2mm}
    \caption{\small Examples of implicit sentence pairs from {\dataset}, extracted from different domains. Unlike explicit relations marked by linking phrases (as shown in Table \ref{table:class_examples}), these pairs contain implicit discourse relations without explicit connective markers.}
  
  %\vspace{-2mm}
  
    \label{table:implicit_sentence_pairs}
    %\vspace{-2mm}
\end{table*}

\subsection{Class-wise Agreement Rates} 
\label{appendix:class_wise_agreement}
The total number of sentence pairs annotated for each class and the agreement rate between the gold label and automatically assigned label are shown in Table \ref{table:class_wise_agreement_table}. As we can see, for the \textsc{Contrasting}, \textsc{Reasoning} and \textsc{Entailment} classes, there is a very high agreement between the human annotated gold label and the automatically annotated label based on distant supervision. This indicates that the annotators possesses a solid understanding of the scientific NLI task. In contrast, the agreement rate for the \textsc{Neutral} class is  low (only $68.3\%$) compared to the $> 93\%$ agreement rates for the other classes. This is because, unlike \textsc{SciNLI} and \textsc{MSciNLI} (where sentence pairs are extracted from full text of the papers), most sentence pairs in {\dataset} are extracted from abstracts of the papers. Given the small number of sentences in paper abstracts, even non-adjacent sentences remain related in many cases resulting in a low agreement for the \textsc{Neutral} class.

\section{Implementation Details}
\label{appendix:implementation_details}
\paragraph{SLM Baselines} We utilize the huggingface\footnote{\url{https://huggingface.co/}} implementations for our SLM baselines in the experiments. For these models, we concatenate the sentence in each pair with a \texttt{[SEP]} token between them and append a \texttt{[CLS]}. We then project the representation for the \texttt{[CLS]} token with a weight matrix $\mathbf{W} \in \mathbb{R}^{d \times 4}$. This projection is then sent as the input to a softmax activation to get the predicted probability distribution over the four classes. 

Each model is fine-tuned for five epochs on different training sets (\textsc{SciNLI}, \textsc{MSciNLI}, \textsc{MSciNLI+}). Early stopping with a patience of 2 epochs is employed while fine-tuning the SLMs. We use the Macro F1 score on the development set of {\dataset} as the early stopping criteria. For all SLM baselines, we use a learning rate of $2e-5$ and a mini-batch size of $64$. We fine-tune the models using the Adam \cite{kingma2014adam} optimizer, and the cross-entropy loss.

\paragraph{LLM Baselines} For open-source LLMs (\textsc{Llama-2}, \textsc{Llama-3}, \textsc{Mistral} and \textsc{Phi-3}), we utilize the Hugging Face library, employing a greedy decoding strategy with no random sampling and a maximum generated token limit of $40$. 
% We also evaluated \textit{Qwen2.5-7B-Instruct} variant of \textsc{Qwen-2.5} \cite{yang2024qwen} using the same zero-shot setup, which achieved $21.38\%$ Macro F1 on the MISMATCHED test set, notably lower than our weakest baseline in Table \ref{table:baseline_results}) (\textsc{Llama-2} at $28.98\%$)—and thus was excluded from the main results to maintain focus on stronger baselines. 
Proprietary models \textsc{GPT-4O} and \textsc{GEMINI-1.5-PRO} were evaluated via their respective official APIs, specifying the model identifiers as "gpt-4o" and "models/gemini-1.5-pro" respectively. For \textsc{GPT-4O}, deterministic output was ensured by setting temperature=$0.0$. For \textsc{GEMINI-1.5-PRO}, default API generation settings were used without specifying temperature or other generation parameters. Our evaluation scripts for both proprietary models incorporated retry logic (up to $3$ attempts upon API failure).

\paragraph{Fine-Tuned LLM Baseline} For the fine-tuned \textsc{Llama-2} experiments (results presented in Appendix \ref{appendix:Fine-Tuned_LLM_Results}), we employed Parameter-Efficient Fine-Tuning (PEFT) using Low-Rank Adaptation (LoRA). The model was fine-tuned specifically on the \textsc{SciNLI} training dataset. Key hyperparameters were configured as follows: \textsc{LoRA} rank (r) was set to $16$ with alpha of $32$, and \textsc{LoRA} dropout was set to $0.05$. The model underwent training for 3 epochs with a learning rate of $2e-3$. We used a per-device batch size of $32$ with $4$ gradient accumulation steps, resulting in an effective batch size of $128$. Training employed the adamw\_bnb\_8bit \cite{dettmers2022eightbit} optimizer with mixed precision (fp16) training. The fine-tuned model was then evaluated on both \textsc{SciNLI} and \textsc{MisMatched} test sets to assess cross-domain performance, with detailed results provided in Table \ref{table:fine-tuned} of Appendix \ref{appendix:Fine-Tuned_LLM_Results}.

\paragraph{Computational Cost.} We fine-tune each SLM baseline using a single NVIDIA RTX A5000 GPU. It takes $\approx 2$ hours to fine-tune each SLM on \textsc{SciNLI} and \textsc{MSciNLI}, and $\approx 4$ hours to fine-tune them on \textsc{MSciNLI+}. For our LLM baselines (\textsc{Llama-2}, \textsc{Llama-3}, \textsc{Mistral} and \textsc{Phi-3}), we utilize one NVIDIA A100-SXM4-80GB GPU. The inference time for all LLMs for {\dataset} is $\approx 0.25$ hours in the zero-shot setting, and $\approx 3.5$ hours in the few-shot (4-shot) setting. The few-shot experiments for \textsc{SciNLI} require $\approx 4$ hours to complete. 

\section{Few-shot Exemplar Selection}
\label{appendix: Few-shot_Exemplar_Selection}
To ensure robust and reliable few-shot performance evaluation, we employed a systematic approach for exemplar selection and ordering across all experiments.

\textsc{Exemplar Selection and Ordering:} For each k-shot experiment, we conducted 3 independent runs to obtain reliable results. In each run, we randomly selected k exemplars (one from each class for balanced representation). The same set of k exemplars was used consistently throughout that entire run for all test examples. The order of exemplars in the prompt was kept identical across all test instances within each run. Final results reported in our tables represent the mean performance and standard deviation computed across these 3 independent runs.

\textsc{MSciNLI+ Exemplar Handling:} Given that \textsc{MSciNLI+} combines \textsc{SciNLI} and \textsc{MSciNLI} datasets, we implemented specific procedures to ensure exemplars truly represent this combined nature. For each independent run on \textsc{MSciNLI+}, we: (1) randomly selected initial candidate exemplars separately from \textsc{SciNLI} and \textsc{MSciNLI} datasets, (2) formed 4-shot prompt combinations from these candidates with the strict requirement that each combination must include at least one exemplar from both original datasets (\textsc{SciNLI} and \textsc{MSciNLI}), and (3) selected three such combinations for our three independent runs. This approach guaranteed that \textsc{MSciNLI+} exemplars always reflected the diverse nature of the combined dataset rather than being dominated by examples from a single source dataset.

\section{Results with Fine-Tuned Llama-2}
\label{appendix:Fine-Tuned_LLM_Results}
We show results of fine-tuned Llama-2 on SciNLI using LoRA. The Macro F1 of this fine-tuned LLM can be seen in Table \ref{table:fine-tuned}.

As we can see, while the performance improves substantially over the prompt based version of the model, there are still differences across the datasets. The in-domain Macro F1 of this model on SciNLI is 83.83\%, which drops to 82.87\% for MisMatched. These results further illustrate the unique linguistic characteristics of the two datasets.

\begin{table}[h]
\centering
\small

  \begin{tabular}{l l}
    \toprule

SciNLI	& MisMatched \\
83.83\%	& 82.87\% \\
\hline

  \end{tabular}
  %\vspace{-3mm}
  \caption{\small Results of Llama-2 fine-tuned on SciNLI.}
    
    \label{table:fine-tuned}

\end{table}

% {\color{blue}\paragraph{Hypothesis-Only Baseline Implementation} For the hypothesis-only baseline experiments (results in Table \ref{table:Hypothesis-only_baseline}), we fine-tuned \textsc{RoBERTa} and \textsc{SciBERT} models using only the hypothesis \textsc{("ONLY 2nd SENTENCE")} from each sentence pair in the \textsc{MSciNLI+} training set. During data preparation, each hypothesis was tokenized independently with model-specific special tokens (\texttt{[CLS]}, \texttt{[SEP]} for BERT-based models) automatically added. Sequences were padded or truncated to a maximum length of $150$ tokens. The models were trained using the same architecture as the full sentence-pair baselines but with single-sentence input. All other hyperparameters remained identical to the main \textsc{SLM baselines}: learning rate $2e-5$, effective batch size $64$ via gradient accumulation), up to $10$ epochs with early stopping (patience=$2$) based on development set Macro F1, Adam optimizer, and cross-entropy loss. This approach tests whether models can achieve reasonable performance using only the hypothesis, without access to premise information.}

\section{Analysis of the “Reasoning” Relation}
\label{reasoning_analysis}

We provide here an in-depth analysis of the “reasoning” relation which is more challenging than the other relations in our {\dataset} dataset. Specifically, we show a confusion matrix between the true labels and the predicted labels by SciBERT (our best performing baseline) on the {\dataset} test set in Table \ref{table:cm}.

\begin{table}[h]
\centering
\small

  \begin{tabular}{l l l l l}
    \toprule

{\bf \diagbox{True}{Predicted}}  &	C	& R	& E	& N \\
\hline
C &	532 & 23 &	30 & 15 \\
R &	60	 &	428 &	79	 &	33 \\
E	 &	55	 &	32	 &	485 &	28 \\
N	 &	71	 &	62	 &	39	 &	428 \\

% SciNLI	& MisMatched \\
% 83.83\%	& 82.87\% \\
\hline

  \end{tabular}
  %\vspace{-3mm}
  \caption{\small Confusion matrix of SciBERT on MisMatched. C: Contrasting; R: Reasoning; E: Entailment; N: Neutral.}
    
    \label{table:cm}

\end{table}

As we can see, the “reasoning” relation is often mistaken with “entailment” by the model. In addition, a fair number of “reasoning” relations are also mistaken as “contrasting” by the model. This results in a lower Macro F1 for the “reasoning” class compared to the other classes.

\section{\textit{Implicit} Relations} 
\label{implicit}
\paragraph{Novelty of Implicit Relations.}
% We would like to emphasize that the knowledge we learn from “implicit” relations is new in our paper and we hope it contributes further to developments in scientific NLI that go beyond the usage of sentence pairs with linking phrases as connectors and can substantially influence the field of natural language inference on the creation of new datasets. 
The “implicit” relations as defined here can help open new directions of research, e.g., to improve discourse coherence analysis by suggesting linking phrases between contiguous sentences for better reading comprehension and natural language understanding.

\paragraph{Examples of Implicit sentence pairs from \dataset}
% \label{appendix:implicit_sentence_pairs}
% The {\dataset} comprises sentence pairs exhibiting implicit discourse relations across multiple domains. 
Table \ref{table:implicit_sentence_pairs} illustrates representative examples from \textsc{Engineering} and \textsc{Psychology} domains, where \textsc{Reasoning} and \textsc{Contrasting} relations must be inferred without explicit connective markers (i.e., without explicit linking phrases between the two sentences).

\paragraph{Further Details on Experimental Setup for \textit{Implicit} Relations} In our experiments with \textit{implicit} relations in Section \ref{section:implicit_relations}, for {\dataset}, \textsc{SciNLI} and \textsc{MSciNLI}, we utilize the \textsc{SciBERT} model fine-tuned as our baseline to predict the labels of the extracted sentence pairs which potentially contain \textit{implicit} relations. However, for predicting the label for the sentence pairs extracted for \textsc{SciNLI}, we fine-tune a separate \textsc{SciBERT} model using \textsc{MSciNLI+} for training and the development set from \textsc{SciNLI} for early stopping. All other implementation details (e.g., learning rate, batch size) are the same as for the SLM baselines.   

After selecting the implicit relations based on the models' (fine-tuned on \textsc{MSciNLI+}) predictions, we fine-tune an out-of-the-box \textsc{SciBERT} model on these selected examples using the same hyper-parameters as for the SLM baselines. %The last checkpoint from this step is further fine-tuned on \textsc{MSciNLI+}. 

The last checkpoint of the model fine-tuned on sentence pairs with \textit{implicit} relations is further fine-tuned on \textsc{MSciNLI+}. Specifically, we initialize the language model layers of \textsc{SciBERT} from the model fine-tuned in the previous step. However, the weight matrix $\mathbf{W} \in \mathbb{R}^{d \times 4}$ (which projects the \texttt{[CLS]} representation to get the probability distribution over the classes) is reinitialized randomly. Furthermore, we use a lower learning rate of $2e-6$ for fine-tuning the models in this step.

%For this step, we do not employ early stopping. 

%we first fine-tune the \textsc{SciBERT} model using \textsc{MSciNLI+} as the training set and the development set from the relevant dataset for each experiment. That is, for our experiments with \textsc{SciNLI}, we use the development set from \textsc{SciNLI} and for  

% \section{Prompt used in LLM Baselines}
% \label{appendix:prompt}
% The prompt that we use for our LLM baselines with \textsc{Llama-2} can be seen in Table \ref{table:prompts}. As we can see, the model is first presented with the two sentences in a sentence pair. Then we ask a multiple-choice question to the model with the class definitions as the choices. We choose this prompt for our experiments because it is reported as the best performing prompt for \textsc{MSciNLI} \cite{sadat-caragea-2024-mscinli}. Note that, the prompt shown in Table \ref{table:prompts} is in a zero-shot format (i.e., without any exemplars). For the few-shot setting, we prepend the example input-output pairs with the prompt. 

% \section{Details on Dataset Release}
% We will make our dataset and code publicly available for academic research purposes on GitHub. 